\documentclass{article}

\usepackage[final]{neurips_2022}

\usepackage{amsmath}
\usepackage{amssymb}
\usepackage{mathtools}
\usepackage{amsthm}

\usepackage{xcolor}

\usepackage{hyperref}
\theoremstyle{plain}
\newtheorem{theorem}{Theorem}[section]
\newtheorem{proposition}[theorem]{Proposition}

\theoremstyle{definition}
\newtheorem{definition}[theorem]{Definition}

\theoremstyle{remark}
\newtheorem{remark}[theorem]{Remark}
\theoremstyle{claim}
\newtheorem{claim}[theorem]{Claim}

\title{A Simple Approach to Automated Spectral Clustering}

\usepackage{amsmath}
\usepackage{amssymb}
\usepackage{algorithm}
\usepackage{algorithmic}
\usepackage{mathrsfs}
\usepackage{multirow}
\usepackage{amsthm}
\usepackage{bm}
\usepackage{scalerel}
\usepackage{makecell}
\usepackage{subfig}

\usepackage{xr}
\usepackage[export]{adjustbox}

\author{%
  Jicong Fan$^{1,2}$,  ~Yiheng Tu$^{3,4}$, ~Zhao Zhang$^{5}$\thanks{Corresponding author}, ~Mingbo Zhao$^{6}$, ~Haijun Zhang$^{7}$\\
  $^1$The Chinese University of Hong Kong, Shenzhen~~ $^2$Shenzhen Research Institute of Big Data\\
  $^3$Chinese Academy of Science, Beijing~~ $^4$University of Chinese Academy of Sciences, Beijing\\
  $^5$Hefei University of Technology, Hefei~~ $^6$Donghua University, Shanghai\\
  $^7$Harbin Institute of Technology, Shenzhen\\
  \texttt{fanjicong@cuhk.edu.cn~~yihengtu@gmail.com~~cszzhang@gmail.com}\\
   \texttt{mzhao4@dhu.edu.cn~~hjzhang@hit.edu.cn}
}

\begin{document}

\maketitle

\begin{abstract}
The performance of spectral clustering heavily relies on the quality of affinity matrix. A variety of affinity-matrix-construction (AMC) methods have been proposed but they have hyperparameters to determine beforehand, which requires strong experience and leads to difficulty in real applications, especially when the inter-cluster similarity is high and/or the dataset is large.  In addition, we often need to choose different AMC methods for different datasets, which still depends on experience. To solve these two challenging problems,  in this paper, we present a simple yet effective method for automated spectral clustering. First, we propose to find the most reliable affinity matrix via grid search or Bayesian optimization among a set of candidates given by different AMC methods with different hyperparameters, where the reliability is quantified by the \textit{relative-eigen-gap} of graph Laplacian introduced in this paper. Second, we propose a fast and accurate AMC method based on least squares representation and thresholding and prove its effectiveness theoretically. 
Finally, we provide a large-scale extension for the automated spectral clustering method, of which the time complexity is linear with the number of data points. Extensive experiments of natural image clustering show that our method is more versatile, accurate, and efficient than baseline methods.
\end{abstract}

\section{Introduction}
Clustering is an important approach to data mining and knowledge discovery.  Particularly,  spectral clustering \citep{790354,shi2000normalized,ng2002spectral,von2007tutorial} has superior performance than k-means clustering \citep{steinhaus1956division},  hierarchical clustering   \citep{johnson1967hierarchical},  DBSCAN \citep{ester1996density}, and mixtures of probabilistic principal component analyzers \citep{tipping1999mixtures} in many applications.  Roughly speaking, spectral clustering consists of two steps: 1) construct an affinity matrix in which each element denotes the similarity between two data points; 2) perform normalized cut \citep{shi2000normalized} on the graph corresponding to the affinity matrix.  K-nearest neighbors (K-NN) and Gaussian kernel 
$k(\bm{x},\bm{y})=\exp(-\Vert \bm{x}-\bm{y}\Vert^2/(2\varsigma^2))$
are two popular methods to construct affinity matrices, where $k$ and $\varsigma$ are hyperparameters.  

As the performance of spectral clustering heavily relies on the quality of affinity matrix,  in recent years,  a variety of methods have been proposed to construct or learn affinity matrices for spectral clustering.  Many of them are in the framework of self-expressive \citep{roweis2000nonlinear,SSC_PAMIN_2013} model, i.e.,
$\mathop{\textup{minimize}}_{\bm{C}}\tfrac{1}{2}\Vert\bm{X}-\bm{X}\bm{C}\Vert_F^2+\lambda\mathcal{R}(\bm{C})$.
Here the columns of $\bm{X}\in\mathbb{R}^{m\times n}$ are the data points drawn from a union of subspaces. $\bm{C}\in\mathbb{R}^{n\times n}$ is a coefficient matrix. $\mathcal{R}(\bm{C})$ denotes a regularization operator on $\bm{C}$. $\lambda$ is a hyperparameter to be determined in advance.
 \citet{SSC_PAMIN_2013} proposed to use $\mathcal{R}(\bm{C})=\Vert\bm{C}\Vert_1:=\sum_{i=1}^n\sum_{j=1}^n\vert c_{ij}\vert$ under a constraint $\text{diag}(\bm{C})=\bm{0}$.  In \citep{SSC_PAMIN_2013}, the affinity matrix for spectral clustering is given by $\bm{A}=\vert\bm{C}\vert+\vert\bm{C}\vert^\top$. The method is called Sparse Subspace Clustering (SSC).  Some theoretical results of SSC can be found in \citep{wang2013noisy,soltanolkotabi2014robust}.
 
Following the self-expressive framework, \citet{LRR_PAMI_2013} let  $\mathcal{R}(\bm{C})=\Vert\bm{C}\Vert_\ast$ (nuclear norm of $\bm{C}$) and proposed a Low-Rank Representation (LRR) method for subspace clustering.
\citet{LSRSCLu2012} and \citet{6836065} used the least squares representation (LSR) model for subspace clustering.  
A few variants of LRR and SSC can be found in \citep{KSSC, li2015structured, LSSLRSC, pmlr-v51-shen16,li2016structured,FAN201736,FAN201839,lu2018subspace,pan2021multi,kang2021structured}.  Recently, deep learning methods were also used to learn affinity matrices for spectral clustering \citep{ji2017deep,zhang2019neural,zhang2019self,lv2021pseudo} and have achieved state-of-the-art performance on many benchmark datasets. 

One common limitation of these spectral or subspace clustering methods is that they have at least one hyperparameter to determine. In the codes of SSC\footnote{\citet{wang2013noisy} and \citet{soltanolkotabi2014robust} provided lower and upper bounds for the $\lambda$ in SSC theoretically, which however depend on the unknown noise level.} and its variants provided by their authors, there is usually one more thresholding parameter for affinity matrix,  which affects the clustering accuracy a lot. In the deep learning clustering methods such as \citep{ji2017deep} and \citep{zhang2019neural}, we need to determine the network structures and regularization parameters, which is much more difficult.  Since clustering is an unsupervised learning problem, the hyperparameters cannot be tuned by cross-validation widely used in supervised learning.  Thus we have to tune the hyperparameters in spectral clustering by experience, which is difficult when the dataset is quite different from those in our experience and/or the inter-class similarity is high compared to the intra-cluster similarity.  Note that SSC, LRR, and their kernel or deep learning extensions have quadratic or even cubic time complexity (per iteration), which further increases the difficulty of hyperparameter selection in clustering large datasets, though there have been a few works improving the computational efficiency \citep{peng2013scalable,cai2014large,wang2014exact,peng2015unified,you2016oracle,
you2016scalable,li2017large,you2018scalable,matsushima2019selective,li2020learnable,chen2020stochastic,kang2020large,fankdd2021,caicvpr2022}.  On the other hand, different datasets often require different AMC methods, which is hard to tackle by experience.


This paper aims at model and hyperparameter selection for spectral clustering and wants to improve the convenience, accuracy,  and efficiency of spectral clustering. Our contributions are as follows.
 \begin{itemize}
\item We propose a \textit{relative-eigen-gap} based automated spectral clustering (AutoSC) method. It finds the Laplacian matrix with largest \textit{relative-eigen-gap} among a set of candidates constructed by different models with different hyperparameters. 
\item We also implement the AutoSC method via Bayesian optimization.  The method can select the possibly best model and optimize the hyperparameters automatically.  Note that any AMC methods (e.g. SSC) can be included in the framework of AutoSC.
\item To improve the accuracy and efficiency of AutoSC, we propose a new AMC method based on least squares representation and thresholding and prove its effectiveness theoretically. 

\item We provide an extension for AutoSC to cluster large-scale datasets.

\end{itemize}
Experiments on seven benchmark image datasets demonstrate the effectiveness of our method. Particularly, our method outperforms state-of-the-art methods of large-scale clustering. 

\section{Related work}

\paragraph{Exploiting eigenvalue information for clustering}  As the number of zero eigenvalues of a Laplacian matrix is equal to the number of connected components of the graph \citep{von2007tutorial}, a few researchers took advantage of eigenvalue information in spectral clustering \citep{regsc2005,regsc2006,Ji_2015_ICCV,hu2017finding,lu2018subspace}.
For instance,  \citet{Ji_2015_ICCV} utilized eigen-gap to determine the rank of the Shape Interaction Matrix. 
But the method requires determining another hyperparameter $\gamma$ beforehand and needs to perform spectral clustering multiple times. The methods of \citep{regsc2005,hu2017finding,lu2018subspace} are based on iterative optimization (need to perform eigenvalue decomposition at every iteration) and hence are not effective in handling large-scale datasets. In addition,  the BDR method of \citep{lu2018subspace} has two hyperparameters ($\lambda$, $\gamma$) to determine by experience, although it outperformed SSC and LRR on some datasets. A comparison is shown in Figure \ref{fig_lambda}.

\paragraph{{Automated machine learning}} Automated model and hyperparameter selection for supervised learning have been extensively studied \citep{hutter2019automated}.  In contrast, the study for unsupervised learning is very limited.  The reason is that in unsupervised learning there is no ground truth or reliable metric to evaluate the performance of algorithms. Concurrently to our work,  \citet{poulakis2020unsupervised} also attempted to do automated clustering. Specifically,  \citet{poulakis2020unsupervised} proposed to use meta-learning to select clustering algorithm and use a heuristic combination of some clustering validity metrics such as Silhouette coefficient \citep{liu2010understanding} and S$\_$Dbw \citep{989517} as an objective to maximize via grid search or Bayesian optimization \citep{jones1998efficient}. One problem is that these metrics are mainly based on Euclidean distance or densities and hence may not be suitable to evaluate the clustering performance of non-distance or non-density based clustering algorithms. Another one is that there is no unified metric to compare different clustering algorithms.

\section{Automated Spectral Clustering (AutoSC)}
\subsection{Preliminary Knowledge}
Let $\bm{A}\in\mathbb{R}^{n\times n}$ be an affinity matrix constructed from a given data matrix $\bm{X}\in\mathbb{R}^{m\times n}$. The corresponding graph is denoted by $G=(V,E)$, where $V=\lbrace v_1,\ldots,v_n\rbrace$ is the vertex set and $E=\lbrace e_1,\ldots,e_l\rbrace$ is the edge set.  The degree matrix of a graph $G$ is defined as
$\bm{D}=\textup{diag}(\bm{A}\bm{1})$, where $\bm{1}=[1,\ldots,1]^\top$. Our goal is to partition the vertices into $k$ disjoint nonempty subsets $C_1,\ldots,C_k$. Let $\mathcal{C}=\lbrace C_1,\ldots,C_k\rbrace$. It is expected to find a partition $\mathcal{C}$ that minimizes the following metric.

\begin{definition}[MNCut]
 The multiway normalized cut (MNCut) \citep{meila2001multicut} is defined as
\begin{equation}
\textup{MNCut}(\mathcal{C})=\sum_{i=1}^k\sum_{j\neq i}\dfrac{\text{Cut}(C_i,C_j)}{\text{Vol}(C_i)},
\end{equation}
where $\text{Cut}(C_i,C_j)=\sum_{u\in C_i}\sum_{v\in C_j} A_{uv}$ and $\text{Vol}(C_i)$ denotes the sum of vertex degrees of $C_i$. 
\end{definition}

The normalized graph Laplacian matrix is defined as
\begin{equation}
\bm{L}=\bm{I}-\bm{D}^{-1/2}\bm{A}\bm{D}^{-1/2},
\end{equation}
where $\bm{I}$ is an identity matrix. The normalized graph Laplacian is often more effective than the unnormalized one in spectral clustering (some theoretical justification was given by \citep{von2007tutorial}).
Let $\sigma_i(\bm{L})$ be the $i$-th smallest eigenvalue of $\bm{L}$. 
The following claim shows the connection between $\textup{MNCut}(\mathcal{C})$ and $\bm{L}$.
\begin{claim}\label{claim_sumk}
The sum of the $k$ smallest singular values of $\bm{L}$ quantifies the potential connectivity among $C_1,\ldots,C_k$:
$\textup{MNCut}(\mathcal{C})\geq \sum_{i=1}^k\sigma_i(\bm{L}).$
\end{claim}
The claim can be easily proved by using Lemma 4 of \citep{meila2001multicut}. We defer all proof of this  paper to the appendices.  Because the multiplicity $k$ of the eigenvalue $0$ of $\bm{L}$ equals the number of connected components in $G$ \citep{von2007tutorial},
we expect to construct an affinity matrix $\bm{A}$ from $\bm{X}$ such that $\bm{L}$ has $k$ zero eigenvalues. Thus the optimal partition means $\textup{MNCut}(\mathcal{C})=\sum_{i=1}^k\sigma_i(\bm{L})=0$.

\subsection{Relative Eigen-Gap Guided Search}\label{sec_reg}

In practice, we may construct an $\bm{A}$ such that $\sum_{i=1}^k\sigma_i(\bm{L})$ is as small as possible because guaranteeing zero eigenvalues is difficult. But this is not enough because $\bm{L}$ may have $k+1$ or more very small or even zero eigenvalues. The second smallest eigenvalue of the Laplacian matrix of a graph $G$ is called the algebraic connectivity of G (denoted by $ac(G)$) \citep{fiedler1973algebraic}.
We have $ac(G)=0$ if and only if $G$ is not connected. When $G$ has $k$ disjointed components, there are $k$ algebraic connectivities, denoted by $ac(C_1),\ldots,ac(C_k)$.  Based on this, we have
\begin{claim}\label{claim_k+1}
The $k$+$1^{th}$ smallest eigenvalue of $\bm{L}$ quantifies the least potential connectivity of partitions {\small $C_1,\ldots,C_k$} of $\mathcal{C}$:
\begin{equation}
\min_{1\leq i\leq k}\textup{MNCut}(C_i)\geq \sigma_{k+1}(\bm{L}).
\end{equation}
\end{claim}
In other words, $\sigma_{k+1}(\bm{L})$ measures the difficulty in segmenting each of $C_i$ into two subsets. Hence, when $\sigma_{k+1}(\bm{L})$ is large, the partitions $C_1,\ldots,C_k$ are stable.
Based on Claim \ref{claim_sumk} and Claim \ref{claim_k+1}, we may construct an $\bm{A}$ that has small $\sum_{i=1}^k\sigma_i(\bm{L})$ and large $\sigma_{k+1}(\bm{L})$ simultaneously, by solving
\begin{equation}\label{eq.opt_problem}
\begin{aligned}
\mathop{\textup{maximize}}_{\theta}&\ \ \sigma_{k+1}(\bm{L})-\dfrac{1}{k}\sum_{i=1}^k\sigma_i(\bm{L}),\\
\textup{subject to}&\ \ \bm{L}=\bm{I}-\bm{D}^{-1/2}\bm{A}\bm{D}^{-1/2},\  \bm{A}=f_{\theta}(\bm{X}).
\end{aligned}
\end{equation}
where $f_{\theta}:\mathbb{R}^{m\times n}\rightarrow \mathbb{R}^{n\times n}$ is a function with parameter $\theta$, e.g. $\bm{A}=[\exp(-\Vert \bm{x}_i-\bm{x}_j\Vert^2/(2\varsigma^2))]-\bm{I}$.
It is difficult to solve \eqref{eq.opt_problem} because of the composition of $f_{\theta}$, symmetric normalized Laplacian, and eigenvalue decomposition.
On the other hand, in \eqref{eq.opt_problem}, we have to choose $f$ in advance, which requires domain expertise or strong experience because different dataset usually needs different $f$. 

Note that different $f$ can result in very different distributions of eigenvalues and the small eigenvalues are sensitive to $f$, $\theta$,  and noise.  Hence the objective in \eqref{eq.opt_problem} is not effective to compared different $f$ and $\theta$. In this paper, we define a new metric \textit{relative-eigen-gap} as follows
\begin{equation}
\textup{reg}(\bm{L}):=\dfrac{\sigma_{k+1}(\bm{L})-\frac{1}{k}\sum_{i=1}^k\sigma_i(\bm{L})}{\frac{1}{k}\sum_{i=1}^k\sigma_i(\bm{L})+\varepsilon},
\end{equation}
where $\varepsilon$ is a small constant (e.g. $10^{-6}$) to avoid zero denominator.  $\textup{reg}(\bm{L})$ is not sensitive to the scale of the small eigenvalues. Therefore,
instead of  \eqref{eq.opt_problem}, we propose to solve
\begin{equation}\label{eq.main_searc}
\begin{aligned}
\mathop{\textup{maximize}}_{(f,\theta)\in\mathcal{F}\times {\Theta}}&\ \ \textup{reg}(\bm{L}),\\
\textup{subject to}&\ \ \bm{L}=\bm{I}-\bm{D}^{-1/2}\bm{A}\bm{D}^{-1/2},\  \bm{A}=f_{\theta}(\bm{X}),
\end{aligned}
\end{equation}
where $\mathcal{F}$ is a set of pre-defined functions and $\Theta$ is a set of hyperparameters.  In fact, \eqref{eq.main_searc} is equivalent to choosing one $\bm{A}$ (or $\bm{L}$) from a set of candidates constructed by different $f$ with different $\theta$, of which the \textit{relative-eigen-gap} is largest. The best $\theta$ can be found using grid search (or even random search). For convenience, we call the method AutoSC-GD.
Table \ref{tab_f_theta} shows a few examples of $f$ and its parameters. One may use a weighted sum of affinity matrices given by different $f$, like \citep{huang2012affinity}, which however will introduce more hyperparameters.
\begin{table}[h]
\centering
\caption{A few examples of $f$ and its $\theta$ for AMC (AASC: \citep{huang2012affinity})}\label{tab_f_theta}
\begin{tabular}{c|c|c|c|c|c|c|c|c}
\hline
$f$ &K-NN & $\epsilon$-neighborhood & Gaussian kernel & SSC & LRR &LSR & KSSC & AASC\\ \hline

$\theta$ & $K$ & $\epsilon$  & $\sigma$ & $\lambda$ & $\lambda$ &$\lambda$  &$\lambda,\sigma$ & $\sigma_1,\sigma_2,\ldots$\\ \hline
\end{tabular}
\end{table}

The following theorem\footnote{This theorem is a modified version of Theorem 1 in \citep{regsc2005}, which is for the eigen-gap $\sigma_{k+1}-\sigma_k$ of $\bm{L}$. Here we consider $\textup{reg}(\bm{L})$ instead.} shows the connection between $\textup{reg}(\bm{L})$ and the stability of the clustering $\mathcal{C}$.
\begin{theorem}\label{the_1}
Let $\mathcal{C}$ and $\mathcal{C}'$ be two partitions of the vertices of $G$, where $\vert\mathcal{C}\vert=\vert\mathcal{C}'\vert=k$.  Define the distance between $\mathcal{C}$ and $\mathcal{C}'$ as
$\textup{dist}(\mathcal{C}, \mathcal{C}')=1-\tfrac{1}{k}\sum_{C_i\in\mathcal{C}}\sum_{C_j'\in\mathcal{C}'}\tfrac{(\textup{Vol}(C_i\cap C_j'))^2}{\textup{Vol}(C_i)\textup{Vol}(C_j')}$.
Suppose $\eta k\varepsilon\geq\sum_{i=1}^k\sigma_i(\bm{L})\geq k\varepsilon$ and $\textup{reg}(\bm{L})>(k-1)\eta/2$. Let $\delta=\max\big(\textup{MNCut}(\mathcal{C})-\sum_{i=1}^k\sigma_i(\bm{L}),\textup{MNCut}(\mathcal{C}')-\sum_{i=1}^k\sigma_i(\bm{L})\big)$.
Then 
\begin{equation}
\textup{dist}(\mathcal{C}, \mathcal{C}')<\tfrac{1.5\delta\varepsilon^{-1}}{\textup{reg}(\bm{L})+(1-k)\eta/2}.
\end{equation}
\end{theorem}
It indicates that when $\textup{reg}(\bm{L})$ is large and $\delta$ is small, the partitions $\mathcal{C}$ and $\mathcal{C}'$ are close to each other. Thus, the clustering has high stability. When $\sum_{i=1}^k\sigma_i(\bm{L})=k\varepsilon$, we have
 $\textup{dist}(\mathcal{C}, \mathcal{C}')<\tfrac{6\delta}{\sigma_{k+1}(\bm{L})-k\varepsilon}$,
which means the larger $\sigma_{k+1}(\bm{L})$ the more stable clustering.

\textbf{Compare $\textup{reg}(\bm{L})$  with \citep{regsc2006}}
It is worth noting that \citet{regsc2006} proposed to minimize
$\text{J}(\bm{L}):=\frac{1}{k}\sum_{i=1}^k\sigma_i(\bm{L})+\alpha \left(\sigma_k(\bm{L})-\sigma_{k+1}(\bm{L})\right)^2$ to find a good affinity matrix for spectral clustering. Although $\textup{reg}(\bm{L})$ and $\text{J}(\bm{L})$ seem similar, they are essentially different. As mentioned before, different AMC methods may lead to different scales for the small eigenvalues, which makes it difficult to compare different AMC methods using $\text{J}(\bm{L})$. In addition, $\text{J}(\bm{L})$ has a hyperparameter $\alpha$ to determine beforehand, which violates our goal of searching models and hyperparameters. A comparative study ($\alpha\leq 0$) is in Section \ref{sec_autosc_exp} (Table \ref{tab_egreg}).

\subsection{AutoSC via Bayesian Optimization}
Bayesian optimization (BO) \citep{jones1998efficient} has become a promising tool for hyperparameter optimization of supervised machine learning algorithms \citep{snoek2012practical,klein2017fast}. Given a black-box function $g:\mathcal{X}\rightarrow\mathbb{R}$, BO aims to find an $\bm{x}^\ast\in\mathcal{X}$ that globally minimizes $g$ and usually has three steps. The first step is finding the most promising point $\bm{x}_{t+1}\in\mathop{\textup{argmax}}_{\bm{x}}a_{p(g)}(\bm{x})$ by numerical optimization, where $a_{p(g)}:\mathcal{X}\rightarrow\mathbb{R}$ is an acquisition function (e.g. Expected Improvement) relying on an prior $p(g)$ (e.g. Gaussian processes \citep{williams2006gaussian}).  The second step is evaluating the expensive and possibly noisy function $y_{t+1}\sim g(\bm{x})+\mathcal{N}(0,\sigma^2)$ and adding the new sample $(\bm{x}_{t+1},y_{t+1})$ to the observation set $\mathcal{D}_t=\{(\bm{x}_{1},y_{1}),\ldots,(\bm{x}_{t},y_{t})\}$. The last step is updating $p(g)$ and $a_{p(g)}$ using $\mathcal{D}_{t+1}$.  

As an alternative to the grid search for \eqref{eq.main_searc},  we can maximize $\textup{reg}(\bm{L})$ via BO.   Suppose we have a set of different AMC models, i.e., $\mathcal{F}=\{f_1,f_2,\ldots,f_M\}$. For $i=1,2,\ldots, M$,  let 
$$g_{i}(\theta^{(i)}):=-\textup{reg}(\bm{L}(f_i(\theta^{(i)}|\mathbb{X}))),$$
where $\theta^{(i)}$ denotes the hyperparameters in $f_i$ and $\mathbb{X}$ denotes the dataset. Then we use BO to find
\begin{equation}
\theta_{\ast}^{(i)}=\mathop{\textup{argmin}}_{\theta^{(i)}\in S^{(i)}}g_{i}(\theta^{(i)}),
\end{equation}
where $S^{(i)}$ denotes the set of constraints. Finally we get the best model with its best hyperparameters
\begin{equation}\label{eq_eggs_bo}
f_{\star}(\theta_\ast^{(\star)}|\mathbb{X}),~~~\textup{where}~ \star=\mathop{\textup{argmin}}_{1\leq i\leq M}g_i(\theta_{\ast}^{(i)}).
\end{equation}

For convenience, we denote the method by AutoSC-BO. Note that $\mathcal{F}$ can include any AMC methods such as those in Table \ref{tab_f_theta} and even DSC \citep{ji2017deep} (see Appendix \ref{appendix_ssc}).

In AutoSC-BO, we use Expected Improvement (EI) acquisition function
\begin{equation}
a_{\textup{EI}}(\bm{s}|\mathcal{D}_t)=\mathbb{E}_p\left[\max(g_{\min}-g(\bm{s}),0)\right],
\end{equation}
where $g_{\min}$ is the best function value known. The closed-form formulation is
\begin{equation}
a_{\textup{EI}}(\bm{s}|\mathcal{D}_t)=(g_{\min}-\mu)\Phi\left(\frac{g_{\min}-{\mu}}{\sigma}\right)+\phi\left( \frac{g_{\min}-{\mu}}{\sigma}\right),
\end{equation}
where $\mu=\mu(\bm{s}|\mathcal{D}_t,\theta_K)$ and $\sigma=\sigma(\bm{s}|\mathcal{D}_t,\theta_K)$ are the mean value and variance of the Gaussian process,  $\phi$ and $\Phi$ are standard Gaussian cumulative density function and probability density function respectively,  and $\theta_K$ denotes the hyperparameters of the Gaussian process.  For the covariance function, we use the automatic relevance determination (ARD) Mat\'{e}rn $5/2$ kernel \citep{matern2013spatial}
\begin{equation}
\begin{aligned}
k_{M52}(\bm{s},\bm{s}')=\theta_0\left(1+\sqrt{5r^2(\bm{s},\bm{s}')}+\frac{5}{3}r^2(\bm{s},\bm{s}')\right)
\times\exp\left(-\sqrt{5r^2(\bm{s},\bm{s}')}\right),
\end{aligned}
\end{equation}
where $r^2(\bm{s},\bm{s}')=\sum_{j=1}^d(s_j-s_j')^2/\theta_j^2$.

\subsection{Discussion on AMC Methods for AutoSC and LSR with Thresholding}
In AutoSC, the size of searching space is $\vert\mathcal{F} \vert\times\prod_{j}\vert\theta_j \vert$, which should be large enough to include effective models and their hyperparameters. A large number of works have shown that the self-expressive models \citep{SSC_PAMIN_2013,LRR_PAMI_2013,lu2018subspace,ji2017deep} often outperform other AMC models such as Gaussian kernel. However, the self-expressive model based AMC methods often require iterative optimization and has at least quadratic time complexity per iteration, which leads to huge time cost in AutoSC.
Although LSR \citep{LSRSCLu2012} has closed-form solution, the clustering accuracy is not satisfactory \citep{lu2018subspace}. In this work, we will show that LSR with a simple post-processing operation can be a good AMC method and can outperform SSC, LRR, and BDR \citep{lu2018subspace}.
Specifically, the LSR model is given as
\begin{equation}\label{eq.LSR_1}
\mathop{\textup{minimize}}_{\bm{C}}\ \ \tfrac{1}{2}\Vert\bm{X}-\bm{X}\bm{C}\Vert_F^2+\tfrac{\lambda}{2}\Vert\bm{C}\Vert_F^2, 
\end{equation}
of which the closed-form solution is
$\bm{C}=(\bm{X}^\top\bm{X}+\lambda\bm{I})^{-1}\bm{X}^\top\bm{X}$.
Let $\textup{diag}(\bm{C})=\bm{0}$ and $\bm{C}\leftarrow\vert \bm{C}\vert$, the affinity matrix can be constructed as $\bm{A}=(\bm{C}+\bm{C}^\top)/2$.
One problem is that the off-diagonal elements of $\bm{A}$ are dense (leading to a connected graph),  which can result in low clustering accuracy. Therefore, we propose to truncate $\bm{C}$ by keeping only the largest $\tau$ elements of each column of $\bm{C}$.  Nevertheless, it is not easy to determine $\tau$ beforehand. When $\tau$ is too small, the corresponding graph will have $k+1$ or more connected components. When $\tau$ is too large, the corresponding graph will have $k-1$ or less connected components. However, $\tau$ can be automatically determined by our AutoSC-GD and AutoSC-BO.

In the case that the data have some low-dimensional nonlinear structures, the similarity between pair-wise columns of $\bm{X}$ cannot be well recognized by the linear regression \eqref{eq.LSR_1}. Therefore, we also consider the following nonlinear regression model
\begin{equation}\label{eq.KLR_0}
\mathop{\textup{minimize}}_{\bm{C}}\ \ \tfrac{1}{2}\Vert\phi(\bm{X})-\phi(\bm{X})\bm{C}\Vert_F^2+\tfrac{\lambda}{2}\Vert\bm{C}\Vert_F^2, 
\end{equation}
where $\phi$ denotes a nonlinear feature map performed on each column of the matrix, i.e. $\phi(\bm{X})=[\phi(\bm{x}_1),\ldots,\phi(\bm{x}_n)]$. In \eqref{eq.KLR_0}, letting $\phi$ be some feature map induced by a kernel function $k(\cdot,\cdot)$ (e.g. polynomial kernel $k(\bm{x}_i,\bm{x}_j)=(\bm{x}_i^\top\bm{x}_j+b)^q$ and Gaussian kernel),  we get the kernel LSR (KLSR):
\begin{equation}\label{eq.KLR_1}
\mathop{\textup{minimize}}_{\bm{C}}\ \tfrac{1}{2}\textup{Tr}\left(\bm{K}-2\bm{K}\bm{C}+\bm{C}^\top\bm{K}\bm{C}\right)+\tfrac{\lambda}{2}\Vert\bm{C}\Vert_F^2, 
\end{equation}
where $\bm{K}=\phi(\bm{X})^\top\phi(\bm{X})$ and $[\bm{K}]_{ij}=k(\bm{x}_i,\bm{x}_j)$. The closed-form solution is
$\bm{C}=(\bm{K}+\lambda\bm{I})^{-1}\bm{K}$.
The post-processing is the same as that for the solution of LSR.

We introduce the following property, a necessary condition of successful subspace clustering, which is similar to the one used in \citep{wang2013noisy,soltanolkotabi2014robust}.
\begin{definition}[Subspace Detection Property]
A symmetric affinity matrix $\bm{A}$ obtained from $\bm{X}$ has subspace detection property if for all $i$,  the nonzero elements of $\bm{a}_i$ correspond to the columns of $\bm{X}$ in the same subspace as $\bm{x}_i$.
\end{definition}

For convenience, let $\pi(i)$ be the index of the subspace $\bm{x}_i$ belongs to and $C_j$ be the index set of the columns of $\bm{X}$ in subspace $j$. We consider the following deterministic model.
\begin{definition}[Deterministic Model]\label{def_determ}
The columns of $\bm{X}\in\mathbb{R}^{m\times n}$ are drawn from a union of $k$ different subspaces and  are further corrupted by noise, where $\textup{dim}(\mathcal{S}_1\cup \cdots \cup\mathcal{S}_k)=d<m\leq n$. Let $\bm{X}=\bm{U}\bm{\Sigma}\bm{V}^\top$ be the SVD of $\bm{X}$,  where $\bm{\Sigma}=\textup{diag}(\sigma_1,\ldots,\sigma_n)$ and $\sigma_1\geq\sigma_2\geq\cdots\sigma_n$.   Let $\gamma=\sigma_{d+1}/{\sigma_d}$.  Denote $\bm{v}_i=(v_{i1},\ldots,v_{in})$ the $i$-th row of $\bm{V}$ and let $\bar{\bm{v}}_i=(v_{i1},\ldots,v_{id})$. Suppose the following conditions hold\footnote{$\gamma$ measures the noise level, $\beta$ is dominated by the difference between subspaces,  and $\mu$ quantifies the incoherence in the singular vectors.}: 1) for every $i\in[n]$,  the $\bar{\tau}$-th largest element of $\{\vert\bar{\bm{v}}_i^\top\bar{\bm{v}}_j\vert:j\in C_{\pi(i)}\}$ is greater than $\alpha$; 2) $\max_{i\in[n]}\max_{j\in [n]\setminus C_{\pi(i)}}\vert\bar{\bm{v}}_i^\top\bar{\bm{v}}_j\vert\leq\beta$; 3) $\max_{i,j,l}\vert v_{il}v_{jl}\vert\leq \mu$.
\end{definition}
Then the following theorem verifies the effectiveness of \eqref{eq.LSR_1} followed by the truncation (thresholding) operation in subspace detection.
\begin{theorem}\label{theorem_ssd}
Suppose $\bm{X}$ is given by Definition \ref{def_determ} and $\bm{C}$ is given by \eqref{eq.LSR_1} with 
\begin{equation}
\tfrac{\left(\rho-\sqrt{\rho^2-4(2\mu d-\Delta)(2\mu m-2\mu d-\Delta)}\right)\sigma_d^2}{4\mu d-2\Delta}<\lambda<\tfrac{\left(\rho+\sqrt{\rho^2-4(2\mu d-\Delta)(2\mu m-2\mu d-\Delta)}\right)\sigma_d^2}{4\mu d-2\Delta}
\end{equation}
where $\rho=2\mu m\gamma^2-\Delta(1+\gamma^2)$ and $\Delta=\alpha-\beta$.
Then the $\bm{C}$ truncated by $\tau\leq \bar{\tau}$ has the subspace detection property.
\end{theorem}

In Theorem \ref{theorem_ssd},  the width of the range of $\lambda$ is $w=\frac{\sqrt{\rho^2-4(2\mu d-\Delta)(2\mu m-2\mu d-\Delta)}\sigma_d^2}{2\mu d-\Delta}$. We see that a larger $\sigma_d$, $\Delta$, or smaller $\gamma$, $d$ leads to a wider range of $\lambda$, which corresponds to a simper clustering problem.  When $\rho^2\leq 4(2\mu d-\Delta)(2\mu m-2\mu d-\Delta)$,  $\lambda$ does not exist. Theorem \ref{theorem_ssd} can be extended to the kernel case \eqref{eq.KLR_1} without the restriction of $d< m$ even when the columns of $\bm{X}$ are drawn from a union of nonlinear low-dimensional manifolds. See Definition \ref{def_kernel_reg}, Definition \ref{def_manifold_dec}, and Theorem \ref{theorem_kernel_poly} in Appendix \ref{more_theory}. Based on Theorem \ref{theorem_ssd} and Theorem \ref{theorem_kernel_poly},  the follow proposition indicates that AutoSC can cluster the data correctly.

\begin{proposition}\label{prop_EGGS_G}
Suppose the affinity matrix $\bm{A}$ given by AutoSC has the subspace or manifold detection property (defined in Appendix \ref{more_theory}) and $\textup{reg}(\bm{L})=\frac{\sigma_{k+1}}{\epsilon}>0$. Then each component of $G$ consists of all columns of $\bm{X}$ in the same subspace or manifold.
\end{proposition}

Now we see that LSR and KLSR with thresholding can provide effective self-expressive affinity matrices for AutoSC without performing iterative optimization. On the other hand, the relative-eigen-gap is able to compare LSR with KLSR, compare different kernels, and evaluate $\lambda$, $\tau$, and kernel parameters. Note that if we use SSC and KSSC instead of LSR and KLSR, AutoSC will be very time-consuming. If we use LSR and KLSR without thresholding, AutoSC may not provide high clustering accuracy. We hope that AutoSC is not only automatic but also accurate and efficient.

\subsection{AutoSC+NSE for Large-Scale Data}

Since the time and space complexity of AutoSC are quadratic with $n$, it cannot be directly applied to large-scale datasets. To solve the problem, we propose to perform Algorithm \ref{alg.AutoSC} on a set of $s$ landmarks of the data (denoted by $\hat{\bm{X}}$) to get a $\hat{\bm{Z}}$. The landmarks can be generated by k-means or randomly. Then we regard $\hat{\bm{Z}}$ as a feature matrix and learn a map $g:\mathbb{R}^m\rightarrow\mathbb{R}^k$ from $\hat{\bm{X}}$ to $\hat{\bm{Z}}$. 
According to the universal approximation theorem \citep{SONODA2017233} of neural networks,  we approximate $g$ by a two-layer neural network and solve
\begin{equation}\label{eq.NSE}
\mathop{\textup{minimize}}_{\bm{W}_1,\bm{W}_2,\bm{b}_1,\bm{b}_2}\ \frac{1}{2s}\Vert \hat{\bm{Z}}-\bm{W}_2\textup{ReLU}(\bm{W}_1\hat{\bm{X}}+\bm{b}_1\bm{1}_s^\top)-\bm{b}_2\bm{1}_s^\top\Vert_F^2
+\frac{\gamma}{2}\left(\Vert\bm{W}_1\Vert_F^2+\Vert\bm{W}_2\Vert_F^2\right),
\end{equation}
where $\bm{W}_1\in\mathbb{R}^{d\times m}$, $\bm{W}_2\in\mathbb{R}^{k\times d}$, $\bm{b}_1\in\mathbb{R}^{d}$, and $\bm{b}_2\in\mathbb{R}^{k}$. Since $\bm{A}$ is sparse, $k$ is often less than $m$, and a neural network is used, we call \eqref{eq.NSE} Neural Sparse Embedding (NSE). We use mini-batch Adam \citep{adam2014} to solve NSE. It is worth noting that NSE is different the method proposed by \citep{li2020learnable}. In \citep{li2020learnable}, the regression is for an affinity matrix, which leads to high computational cost.
The network learned from \eqref{eq.NSE} is applied to $\bm{X}$ to extract a $k$-dimensional feature matrix $\bm{Z}$:
\begin{equation}\label{eq.NSE_Z}
\bm{Z}=\hat{g}(\bm{X})=\bm{W}_2\textup{ReLU}(\bm{W}_1\bm{X}+\bm{b}_1\bm{1}_n^\top)+\bm{b}_2\bm{1}_n^\top.
\end{equation}

Finally, we perform k-means on $\bm{Z}$ to get the clusters.  The procedures are summarized into Algorithm \ref{alg.NSE} (see Appendix \ref{sec_large}). Note that the time complexity of AutoSC+NSE is $O(dmn+\tilde{d}s^2)$, where $\tilde{d}$ depends on the specific AMC method. When $s\ll n$, the time complexity of AutoSC+NSE is linear with the number of data points $n$. Proposition \ref{prop_nn} in Appendix \ref{sec_prop_nse} shows that a small number of hidden nodes in NSE are sufficient to make the clustering succeed.

\section{Experiments}\label{sec_exp}
We test our AutoSC on Extended Yale B Face \citep{Dataset_ExtendYaleB}, ORL Face \citep{ORL_face}, COIL20 \citep{Dataset_COIL20}, AR Face \citep{ARfacedata}, MNIST \citep{lecun1998gradient}, Fashion-MNIST \citep{xiao2017fmnist}, GTSRB \citep{stallkamp2012man}, subsets and extracted features of MNIST and Fashion-MNIST. The descriptions for the datasets are in Appendix \ref{sec_datasets}. Our MATLAB codes are available at \url{https://github.com/jicongfan/Automated-Spectral-Clustering}.

\subsection{Intuitive validation of AutoSC}\label{sec_autosc_exp}
\begin{figure}[h!]
\vspace{-10pt}
\subfloat[Clustering accuracy and $\textup{reg}(\bm{L})$ (rescaled between 0 and 1) and clustering accuracy of LSR and KLSR on the first 5 subjects of the Extended Yale Face B database.]{\includegraphics[width=7cm,valign=t]{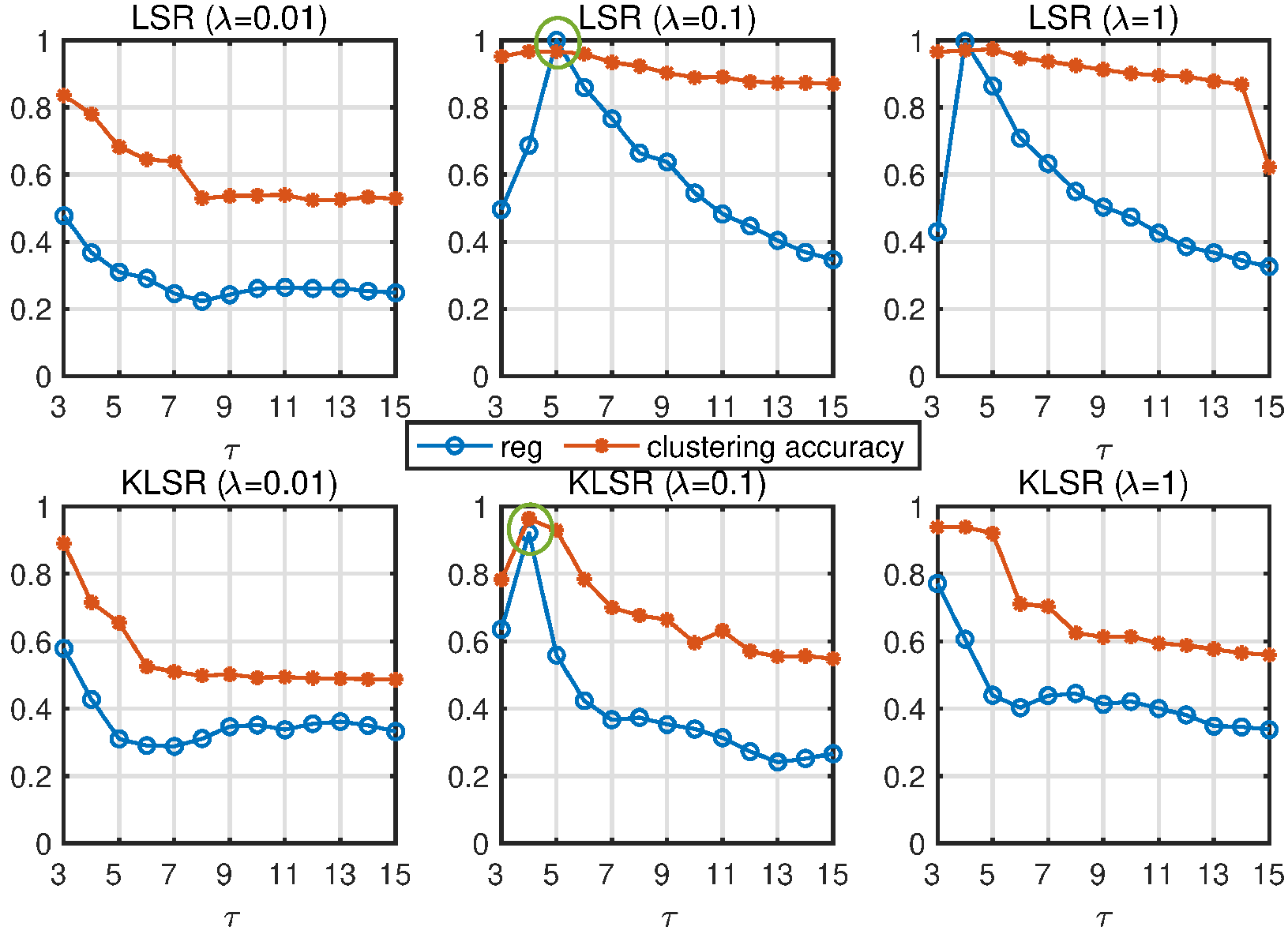}}
\hfil
\subfloat[Clustering accuracies of SSC, LRR, and BDR-B with different hyperparameter $\lambda$ in comparison to AutoSC-GD with LSR and KLSR on the Extended Yale Face B database. The value of $\lambda$ used in BDR-B has been divided by 10. The $\gamma$ in BDR-B is chosen from $\lbrace 0.01,0.1,1\rbrace$ and the best one is used for each $\lambda$. In Case (b), the time costs of SSC, LRR, BDR-B, and AutoSC-GD are 9.5s, 33.0s, 7.6s, and 1.6s respectively. ]{\includegraphics[width=6.5cm,valign=t]{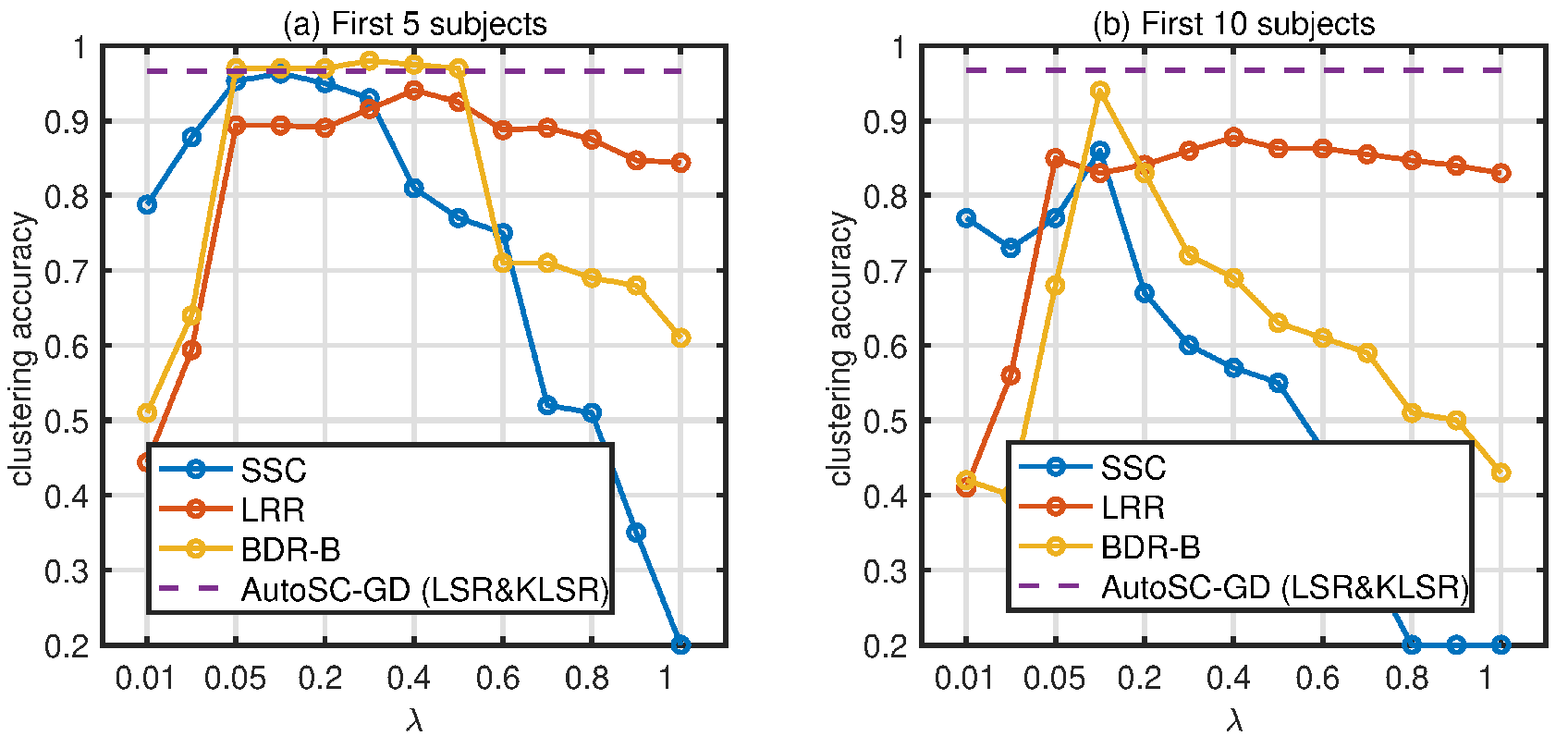}}
\caption{Examples about $\textup{reg}(\bm{L})$, clustering accuracy, and hyperparameters on Extended Yale B.} \label{fig_lambda}
\end{figure}

\paragraph{Performance of Relative-Eigen-Gap}
First, we use LSR and KLSR to show the effectiveness of the proposed $\textup{reg}(\bm{L})$. Figure \ref{fig_lambda}(i) shows an intuitive example of the performance of LSR and KLSR in clustering a subset of the Extended Yale B database,  where for \eqref{eq.KLR_1} we use Gaussian  kernel with $\varsigma=\frac{1}{n^2}\sum_{ij}\Vert\bm{x}_i-\bm{x}_j\Vert$.  We see that: 1) in LSR and KLSR, for a fixed $\lambda$ (or $\tau$), the $\tau$ (or $\lambda$) with larger $\textup{reg}(\bm{L})$ provides higher clustering accuracy; 2) for a fixed $\lambda$ and a fixed $\tau$, if LSR has a  larger $\textup{reg}(\bm{L})$, its clustering accuracy is higher than that of KLSR, and vice versa.  We conclude that roughly a larger $\textup{reg}(\bm{L})$ indeed leads to a higher clustering accuracy, which is consistent with our theoretical analysis in Section \ref{sec_reg}.

\renewcommand\thesubfigure{\roman{subfigure}}

Now we show the superiority of LSR and KLSR compared to a few important AMC methods.
Figure \ref{fig_lambda}(ii) shows the clustering accuracy of SSC \citep{SSC_PAMIN_2013}, LRR \citep{LRR_PAMI_2013}, and BDR-B \citep{lu2018subspace} with different hyperparameters and our method Auto-GD with LSR and KLSR (detailed by Algorithm \ref{alg.AutoSC} in the supplementary material) on the Extended Yale Face B subset. 
SSC and BDR-B are sensitive to the value of $\lambda$, especially for the relatively difficult task, say Figure \ref{fig_lambda}(ii-b). LRR is not sensitive to the value of $\lambda$ but its accuracy is low. LSR and KLSR are more accurate and efficient than other methods.

\begin{figure}[h!]
\subfloat[$\text{reg}(\bm{L})$ and clustering accuracy of KLSR and KSSC in each iteration of BO.]
{\includegraphics[width=6.7cm]{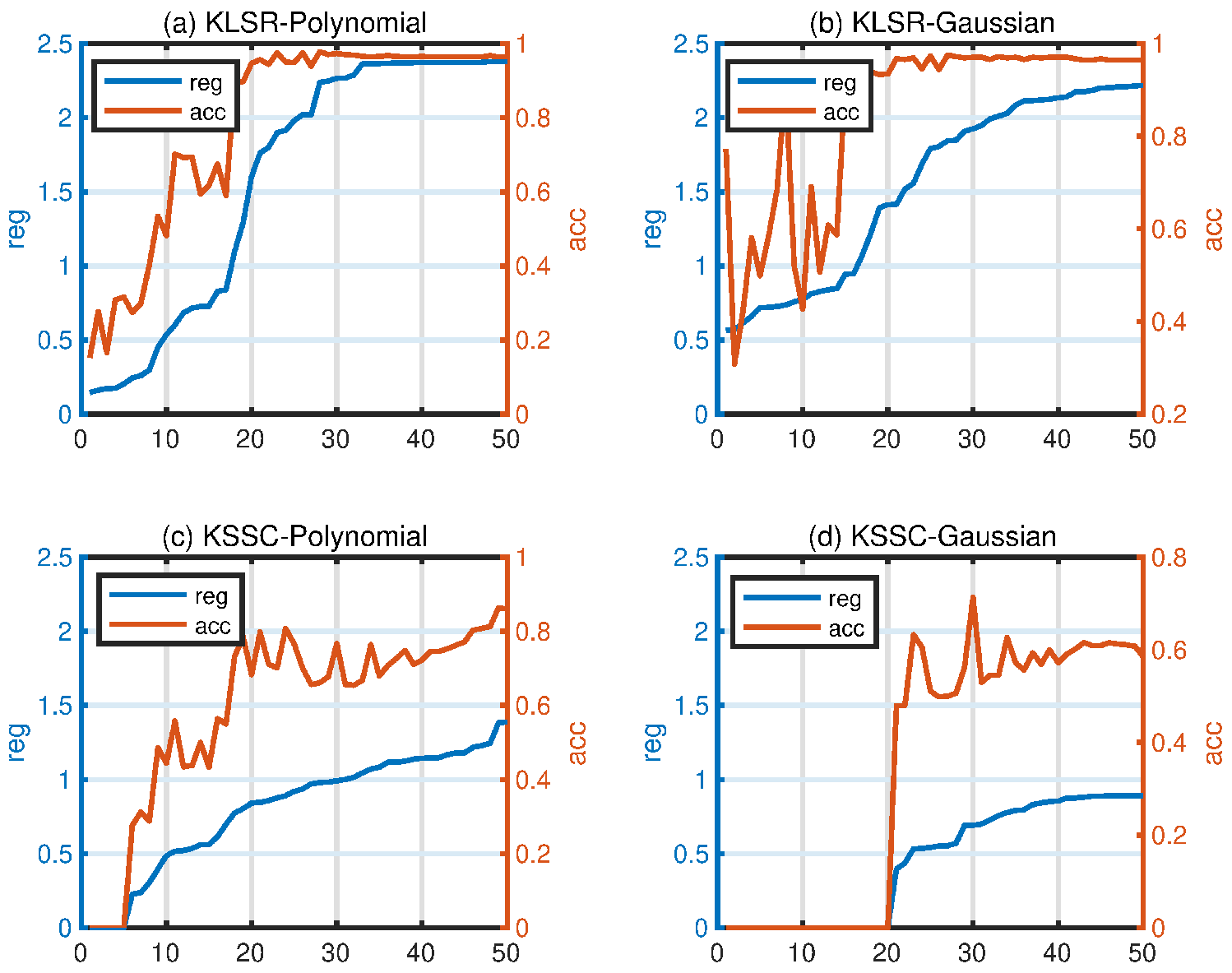}}
\hfil
\subfloat[$\text{reg}(\bm{L})$ and hyperparameters of KLSR and KSSC in each iteration of BO.]
{\includegraphics[width=6.7cm]{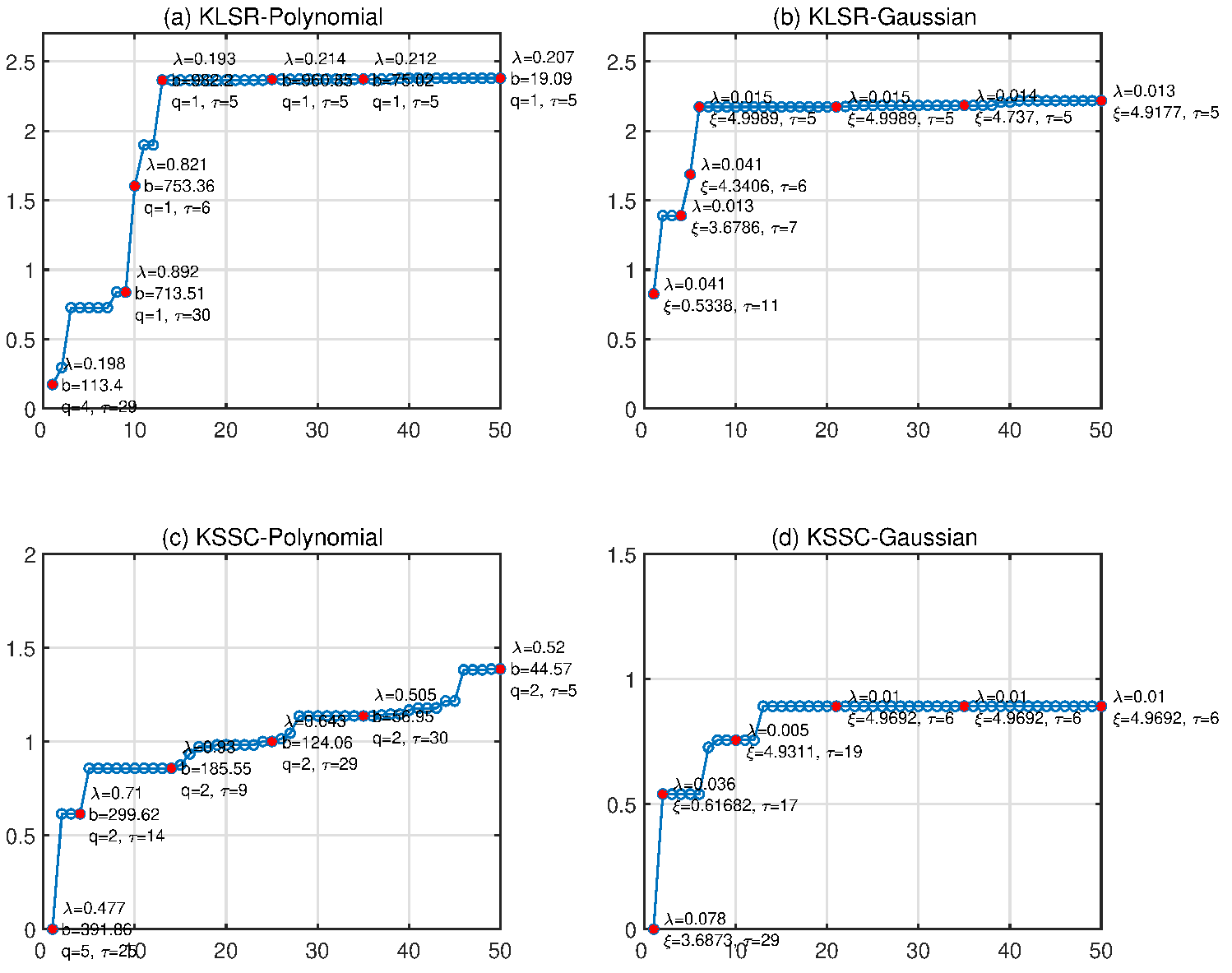}}
\caption{AutoSC-BO with KSSC and KLSR on the first 10 subjects of YaleB Face dataset.}\label{fig_yaleb10_ssc_par}
\end{figure}

\paragraph{The performance of AutoSC-BO}
We show the performance of AutoSC-BO with many AMC methods such as SSC \citep{SSC_PAMIN_2013} and LSR. Taking the KLSR model \eqref{eq.KLR_1} with polynomial kernel as an example, the parameters are $\theta=(\lambda,b,q,\tau)^\top$ and the constraints are given by $S=\{\lambda\in\mathbb{R}:\lambda_{min}\leq\lambda\leq \lambda_{max}; ~b\in\mathbb{R}: b_{min}\leq b\leq b_{max};~q\in\mathbb{Z}^{+}: q_{min}\leq q\leq q_{max};~\tau\in\mathbb{Z}^{+}: \tau_{min}\leq \tau\leq \tau_{max}\}$.  More details are in Appendix \ref{appendix_ssc}. Shown in Table \ref{bo_amc}, larger reg corresponds to higher clustering accuracy and KLSR with polynomial kernel (the optimal $q$ is 1) performs best. Figure \ref{fig_yaleb10_ssc_par} shows the performance of KSSC \citep{KSSC} and KLSR in each iteration of AutoSC-BO.

\begin{table}[h]
\centering
\caption{Clustering accuracy of AutoSC-BO with many methods on Yale Face B dataset (first 10 subjects). All hyperparameters of the kernel functions were optimized via Bayesian optimization.}\label{bo_amc}
\begin{footnotesize}
\begin{tabular}{c|cccccccc}\hline
 AMC & \makecell{$\epsilon$-neigh\\-borhood}& \makecell{Polynomial\\kernel}  & \makecell{Gaussian\\kernel}&   \makecell{KSSC \\(Gauss)} & \makecell{KSSC\\ (Poly)} & \makecell{KLSR\\ (Gauss)}& \makecell{KLSR\\(Poly)}\\ \hline
reg$_\text{max}$ & 0.776 &1.294 &1.307  &0.892 &1.388  &2.217  &2.379\\
Accuracy & 0.325 &0.389 &0.393 &0.584 & 0.859  & 0.963 & 0.966\\ \hline
\end{tabular}
\end{footnotesize}
\vspace{-10pt}
\end{table}

\paragraph{Relative-Eigen-Gap versus Eigen-Gap} 
We compare the proposed relative-eigen-gap (reg$(\mathbf{L})$) with eigen-gap (denoted by eg$(\mathbf{L})$), and the regularizer J$(\mathbf{L},\alpha)$ proposed by \citep{regsc2006} with different $\alpha$. Note that in AutoSC, we need to minimize J$(\mathbf{L},\alpha)$ instead. The clustering results of AutoSC-GD are reported in Table \ref{tab_egreg} (details about the datasets are in Section \ref{sec_exp}). We see that our reg$(\mathbf{L})$ outperforms eg$(\mathbf{L})$ and J$(\mathbf{L},\alpha)$ in all cases except AR.
\begin{table}[h!]
\centering
\caption{The comparison of AutoSC-GD with reg$(\mathbf{L})$, eg$(\mathbf{L})$, and J$(\mathbf{L},\alpha)$}\label{tab_egreg}
\begin{footnotesize}
\begin{tabular}{l|cccccc}
\hline
& {YaleB} &{ORL}&{ COIL20} &AR &{ MNIST}&{ F-MNIST}\\ \hline
eg$(\mathbf{L})$ &{0.790}&{0.768} &{0.619}& 0.805&0.663&0.562\\
J$(\mathbf{L},0$) &{0.823}&\textbf{0.795}&0.750&0.804&0.726& 0.516\\
J$(\mathbf{L},-0.1$) &{0.818}&0.788&0.768&0.826&0.718&0.519\\
J$(\mathbf{L},-1$) &{0.812}&0.785 &0.769&0.817&0.722&0.523\\
J$(\mathbf{L},-10$) &{0.804}& 0.788&0.768& \textbf{0.832}&0.731&0.539\\
J$(\mathbf{L},-100$) &{0.788}& 0.765& 0.769 &0.817&0.735&0.545\\
reg$(\mathbf{L})$ &\textbf{0.897}&\textbf{0.795}&\textbf{0.782}&{0.786} &\textbf{0.755}&\textbf{0.595}\\ \hline
\end{tabular}
\end{footnotesize}
\vspace{-10pt}
\end{table}

\subsection{Comparative studies of AutoSC and baselines}
First we compare AutoSC with SSC, LRR \citep{LRR_PAMI_2013}, LSR \citep{LSRSCLu2012}, EDSC \citep{6836065}, KSSC, SSC-OMP \citep{you2016scalable}, BDR-Z \citep{lu2018subspace}, and BDR-B \citep{lu2018subspace} on six smaller datasets.  
The clustering accuracy and time cost are reported in Table \ref{tab_five}. AutoSC-GD and AutoSC-BO outperformed other methods significantly in almost all cases. SSC-OMP and AutoSC-GD are more efficient than SSC, LRR, EDSC, and KSSC. The time cost of AutoSC-BO is much higher than that of AutoSC-GD because the former optimizes all hyperparameters including $\lambda$, $\tau$, and kernel parameters (e.g. $b,q$).

\renewcommand\tabcolsep{2.5pt}
\begin{table*}[h]
\centering
\caption{Clustering performance on the six small datasets. For the MNIST-1k and Fahsion-MNIST-1k, we report the average results of 20 trials because the subset is formed randomly. AutoSC chose LSR for Yale B and AR and chose KLSR for others datasets.  The NMI results are in Table \ref{tab_five_nmi} (See Appendix \ref{sec_nmi}).}\label{tab_five}
\begin{footnotesize}
\begin{tabular}{c|c|cccccccccc}\hline
& & SSC & LRR &LSR&  EDSC & KSSC &  {\scriptsize SSC-OMP}&BDR-Z	&BDR-B	&{\scriptsize AutoSC-GD}&{\scriptsize AutoSC-BO}\\ \hline
\multirow{2}{*}{Yale B}& acc&0.723 &	0.643 &0.592	& 0.806 &0.649	&0.768	&0.596	&0.719	&\textbf{0.897}& \textbf{0.909}\\
& time&273.8	&928.1&	7.3 & 58.6 &464.3 &	\textbf{8.9}	&368.8	&368.8	&19.1& 78.3 \\ \hline
\multirow{2}{*}{ORL}&acc&0.711 &	0.762&	0.680& 0.712&0.707	&0.665	&0.739	&0.735 &\textbf{0.795}& \textbf{0.803}\\
&time &2.7	&8.8	& \textbf{0.5} & 2.0 &2.6	&\textbf{0.4}&3.9	&3.9	&2.3& 20.5         \\ \hline
\multirow{2}{*}{COIL20}&acc&0.871&	0.729&0.695 &0.759	&\textbf{0.912}&	0.658	&0.713&	0.791&	0.782 & \textbf{0.878}\\
&time& 61.8	&221.2	&\textbf{1.4}&15.4 &100.6	&\textbf{2.5}&86.8	&86.8	&7.6& 39.2 \\ \hline
\multirow{2}{*}{AR}&acc&0.718&	0.769& 0.665 & 0.673	&0.726&	{0.669}&	0.745&	0.751&	{\textbf{0.786}}& \textbf{0.826}\\
&time &317.5	&1220.6& \textbf{14.5} &	69.1 & 627.4	&57.6&	578.7	&578.7&	\textbf{43.4}&130.6 \\ \hline
\multirow{2}{*}{{ \makecell{MNIST\\-1k}}}&acc&\makecell{0.596\\(0.054)}	&\makecell{0.513\\(0.037)}	& \makecell{0.554\\(0.041)}& \makecell{0.536\\(0.035)}&\makecell{0.577\\(0.053)}	&\makecell{0.542\\(0.038)}&	\makecell{0.576\\(0.037)}	&\makecell{0.578\\(0.043)}	&\makecell{\textbf{0.615}\\(0.041)}& \makecell{\textbf{0.619}\\(0.038)}\\
&time&24.9&	69.1& \textbf{0.8} &5.2 	&32.8	&\textbf{1.3}	&26.9	&26.9	&2.5& 21.7        \\ \hline
\multirow{2}{*}{\makecell{Fashion-\\MNIST-1k}}&acc& \makecell{0.553\\(0.025)}  & \makecell{0.515\\(0.014)}&\makecell{0.563\\(0.023)}&\makecell{0.544\\(0.017)} &   \makecell{0.548\\(0.016)}    &\makecell{0.566\\(0.034)}  &  \makecell{{0.574}\\(0.019)}  &  \makecell{0.563\\(0.031)}    &\makecell{\textbf{0.581}\\(0.025)}&\makecell{\textbf{0.584}\\(0.021)}\\
&time &    24.1  & 68.5  &\textbf{0.7} & 5.1 &35.9 &   \textbf{1.2}   &25.7    &    25.7  & {2.6}    &22.7 \\
\hline
\end{tabular}
\end{footnotesize}
\end{table*}



We compare our method with LSR, LSC-K \citep{chen2011large}, SSC-OMP \citep{you2016scalable}, and S$^5$C \citep{matsushima2019selective}, and S$^3$COMP-C \citep{chen2020stochastic} on the larger datasets.  The parameter settings are in Appendix \ref{sec_hyper_large}.
Table \ref{tab_large} shows the clustering accuracy and standard deviation of 10 repeated trials on the raw-pixel data of MNIST and Fashion-MNIST. Table \ref{tab_large_fea} shows the results on MNIST and Fashion-MNIST features and GTSRB. Our methods have the highest clustering accuracy in every case. Note that if NSE is ablated,  the accuracy of AutoSC-GD on MNIST(features)-10k, -20k, and -30k are 0.9772,  0.9783 and 0.9864 respectively,  higher than those of AutoSC-GD+NSE, though the time costs increased.

It is worth mentioning that, to the best of our knowledge, the deep clustering method of \citep{zhang2019self} has SOTA performance on Yale B (acc=0.98) and ORL (acc=0.89), the method proposed by \citep{zhang2019neural} has SOTA performance on Fashion-MNIST (acc=0.72), the method proposed by \citep{mahon2021selective} has SOTA performance on MNIST (acc=0.99), and our method has SOTA performance on GTSRB. Nevertheless, we focus on automated spectral clustering.

\renewcommand\tabcolsep{3.8pt}
\begin{table*}[h]
\centering
\caption{Clustering accuracy and time cost (second) on MNIST and Fashion MNIST.  ``---" means the computation is out of memory.}\label{tab_large}
\begin{scriptsize}
\begin{tabular}{ccccccccc}\hline
&  & LSR& LSC-K& SSC-OMP	& S$^5$C & S$^3$COMP-C &AutoSC-GD+NSE& AutoSC-BO+NSE \\ \hline
\multirow{2}{*}{MNIST-10k} & acc& 0.583(0.007)  &{0.652}(0.037)&  0.431(0.014)	& 0.646(0.045) & 0.623(0.028) &\textbf{0.687}(0.035)& \textbf{0.679}(0.034)\\
& time&  154.9  &\textbf{{18.9}} & 26.4& 82.3 & 710.4/20 &\textbf{16.2}&48.3\\ \hline
\multirow{2}{*}{MNIST}  &acc & ---&{0.665}(0.021)&  0.453(0.017)	& 0.627(0.025) & --- &\textbf{0.755}(0.022)& \textbf{0.750}(0.009)\\
& time& --- &329.2& 1178.3	& 961.5 & --- &\textbf{86.9}&\textbf{123.6}\\ \hline
\multirow{2}{*}{\makecell{Fashion-\\MNIST-10k}} & acc& 0.561(0.008)  &{0.571}(0.025)	&0.509(0.038)	&0.565(0.021)&0.569(0.024)&\textbf{0.576}(0.011)& \textbf{0.572}(0.019)\\
& time&  153.6 &\textbf{{18.6}}		&26.8 &107.3 &707.2/20	&\textbf{17.3}&50.9\\ \hline
\multirow{2}{*}{\makecell{Fashion-\\MNIST}} & acc&---& {0.561}(0.015)	&0.359(0.017)	&0.559(0.013)&--- &\textbf{0.586}(0.008)& \textbf{0.578}(0.012)\\
& time& ---& 335.1		&1156.6	&932.6 &--- &\textbf{88.7}&\textbf{122.8}\\ \hline
\end{tabular}
\end{scriptsize}
\end{table*}

\begin{table*}[h!]
\centering
\caption{Clustering accuracy (mean value and standard deviation) and time cost (second) on MNIST and Fashion-MNIST with feature extraction.  ``\slash" means the algorithm was performed on a computational platform not comparable to ours. The underlined values are from \citep{chen2020stochastic}.}\label{tab_large_fea}
\begin{scriptsize}
\begin{tabular}{ccccccccc}\hline
&  & LSC-K&{SSC-OMP} &S$^5$C& S$^3$COMP-C&AutoSC-GD+NSE& AutoSC-BO+NSE \\ \hline
\multirow{2}{*}{MNIST} & acc&0.8659(0.0215)&\underline{0.8159}	&0.7829(0.0283)	&\underline{0.9632} &\textbf{0.9775}(0.0034)& \textbf{0.9741}(0.0044)\\
& time&273.6&\underline{280.6}	&907.5		&\underline{416.8} &\textbf{59.2}&\textbf{115.3}\\ \hline
\multirow{2}{*}{{\makecell{Fashion-\\MNIST}}} & acc&0.6131(0.0298)& 0.3796(0.0217)&0.6057(0.0227)	& --- &\textbf{0.6398}(0.0133)& \textbf{0.6461}(0.0104)\\
& time&251.8 &1013.9&913.2& --- &\textbf{61.9}& \textbf{112.6} \\ \hline

\multirow{2}{*}{{\makecell{GTSRB}}} & acc&0.8711(0.0510)& \underline{0.8252}&0.9044(0.0267)	& \underline{0.9554} &\textbf{0.9873}(0.0126)&\textbf{0.9881}(0.0078)\\
& time&\textbf{31.2}&\slash &98.7& \slash &\textbf{16.8}&69.4\\ \hline
\end{tabular}
\end{scriptsize}
\end{table*}

\section{Conclusion}
We have proposed an automated spectral clustering method. Extensive experiments showed the effectiveness and superiority of our methods over baseline methods. 
The efficiency improvement is from the closed-form solutions of the least squares regressions. The accuracy improvement is from the effectiveness of LSR and KLSR with thresholding and the automation of model and hyperparameter selection. One limitation of this work is that we only considered automated spectral clustering while there are many other clustering methods (e.g. \citep{fankdd2021}) not relying on affinity matrix.

\section*{Acknowledgments}

The work of Jicong Fan was supported in part by the research funding T00120210002 of Shenzhen Research Institute of Big Data and the Youth program 62106211 of the National Natural Science Foundation of China. 
The work of Yiheng Tu was supported in part by the National Natural Science Foundation of China under Grant no. 32171078. 
The work of Zhao Zhang was supported in part by the National Natural Science Foundation of China under Grant no. 62072151 and Anhui Provincial Natural Science Fund for the Distinguished Young Scholars (2008085J30). 
The work of Mingbo Zhao was supported in part by the National Natural Science Foundation of China under Grant no. 61971121.
The work of Haijun Zhang was supported in part by the National Natural Science Foundation of China under Grant no. 61972112 and no. 61832004, the Guangdong Basic and Applied Basic Research Foundation under Grant no. 2021B1515020088. 

The authors appreciate the reviewers' comments and time.

\bibliography{Ref_SC}
\bibliographystyle{named}

\section*{Checklist}

The checklist follows the references.  Please
read the checklist guidelines carefully for information on how to answer these
questions.  For each question, change the default \answerTODO{} to \answerYes{},
\answerNo{}, or \answerNA{}.  You are strongly encouraged to include a {\bf
justification to your answer}, either by referencing the appropriate section of
your paper or providing a brief inline description.  For example:
\begin{itemize}
  \item Did you include the license to the code and datasets? \answerYes{See Section.}
  \item Did you include the license to the code and datasets? \answerNo{The code and the data are proprietary.}
  \item Did you include the license to the code and datasets? \answerNA{}
\end{itemize}
Please do not modify the questions and only use the provided macros for your
answers.  Note that the Checklist section does not count towards the page
limit.  In your paper, please delete this instructions block and only keep the
Checklist section heading above along with the questions/answers below.

\begin{enumerate}
\item For all authors...
\begin{enumerate}
  \item Do the main claims made in the abstract and introduction accurately reflect the paper's contributions and scope?
    \answerYes{}
  \item Did you describe the limitations of your work?
    \answerNA{}
  \item Did you discuss any potential negative societal impacts of your work?
    \answerNA{}
  \item Have you read the ethics review guidelines and ensured that your paper conforms to them?
    \answerYes{}
\end{enumerate}

\item If you are including theoretical results...
\begin{enumerate}
  \item Did you state the full set of assumptions of all theoretical results?
    \answerYes{}
        \item Did you include complete proofs of all theoretical results?
    \answerYes{}
\end{enumerate}

\item If you ran experiments...
\begin{enumerate}
  \item Did you include the code, data, and instructions needed to reproduce the main experimental results (either in the supplemental material or as a URL)?
    \answerYes{}
  \item Did you specify all the training details (e.g., data splits, hyperparameters, how they were chosen)?
    \answerYes{}
        \item Did you report error bars (e.g., with respect to the random seed after running experiments multiple times)?
    \answerYes{}
        \item Did you include the total amount of compute and the type of resources used (e.g., type of GPUs, internal cluster, or cloud provider)?
    \answerYes{}
\end{enumerate}

\item If you are using existing assets (e.g., code, data, models) or curating/releasing new assets...
\begin{enumerate}
  \item If your work uses existing assets, did you cite the creators?
    \answerYes{}
  \item Did you mention the license of the assets?
    \answerNA{}
  \item Did you include any new assets either in the supplemental material or as a URL?
    \answerNA{}
  \item Did you discuss whether and how consent was obtained from people whose data you're using/curating?
    \answerNA{}
  \item Did you discuss whether the data you are using/curating contains personally identifiable information or offensive content?
    \answerNA{}
\end{enumerate}

\item If you used crowdsourcing or conducted research with human subjects...
\begin{enumerate}
  \item Did you include the full text of instructions given to participants and screenshots, if applicable?
    \answerNA{}
  \item Did you describe any potential participant risks, with links to Institutional Review Board (IRB) approvals, if applicable?
    \answerNA{}
  \item Did you include the estimated hourly wage paid to participants and the total amount spent on participant compensation?
    \answerNA{}
\end{enumerate}

\end{enumerate}

\appendix

\section{More discussion about LSR and KLSR}
Note that if $n\ll m$, using the \textit{push-through identity} \citep{henderson1981deriving}, we reformulate $\bm{C}=(\bm{X}^\top\bm{X}+\lambda\bm{I})^{-1}\bm{X}^\top\bm{X}$ as
$\bm{C}=\bm{X}^\top(\lambda\bm{I}+\bm{X}\bm{X}^\top)^{-1}\bm{X}$
to reduce the computational cost from $O(n^3)$ to $O(mn^2)$. In $\bm{C}=(\bm{K}+\lambda\bm{I})^{-1}\bm{K}$, when $n$ is large (e.g.$>5000$), we perform randomized SVD \citep{randomsvd} on $\bm{K}$: $\bm{K}\approx\bm{V}_r\bm{\Sigma}_r\bm{V}_r^\top$. Then
$\bm{C}\approx \bm{V}_r\bm{\Sigma}_r^{1/2}(\lambda\bm{I}+\bm{\Sigma})^{-1}\bm{\Sigma}_r^{1/2}\bm{V}_r^\top$,
where $r=20k$ works well in practical applications. The time complexity of computing $\bm{C}$ is $O(r\tau n+rn^2)$. The computation of the smallest $k+1$ eigenvalues of $\bm{L}$ is equivalent to compute the largest $k+1$ eigenvalues and eigenvectors of $\bm{D}^{-1/2}\bm{A}\bm{D}^{-1/2}$, which is sparse.  The time complexity is $O(k\tau n)$.
We have the following result.
\begin{proposition}\label{prop_kernel}
Let $\hat{\bm{c}}$ be the optimal solution of 
$\mathop{\textup{minimize}}_{\bm{c}}\tfrac{1}{2}\Vert\phi(\bm{y})-\phi(\bm{X})\bm{c}\Vert^2+\tfrac{\lambda}{2}\Vert\bm{c}\Vert^2$,
where $\phi$ is induced by Gaussian kernel and $\bm{y}$ is arbitrary.
Then 
$\Vert\hat{c}_i-\hat{c}_j\Vert\leq \sqrt{2-2\exp\left(-\Vert\bm{x}_i-\bm{x}_j\Vert^2/(2\varsigma^2)\right)}$.
\end{proposition}
It shows that when two data points in $\bm{X}$, e.g. $\bm{x}_i$ and $\bm{x}_j$, are close to each other, the corresponding two elements in $\hat{\bm{c}}$, e.g. $\hat{c}_i$ and $\hat{c}_j$, have small difference.  Hence KLSR with Gaussian kernel utilizes  local information to enhance $\bm{C}$.

In LSR and KLSR, let $\lambda\in{\Lambda}$, $\tau\in\mathcal{T}$, and $\Theta={\Lambda}\times \mathcal{T}$. The algorithm of AutoSC-GD with only LSR and KLSR is shown in Algorithm \ref{alg.AutoSC}. The total time complexity is 
$$O\left(\vert \Lambda\vert(mn^2+r\bar{\tau} n+rn^2)+2\vert \Lambda\vert\vert \mathcal{T}\vert k\bar{\tau} n\right),$$ where $\bar{\tau}$ denotes the mean value in $\mathcal{T}$. The time complexity is at most $O(\vert \Lambda\vert\left(mn^2+\vert \mathcal{T}\vert kmn)\right)$ when $\tau\leq r\leq m\leq n$.
It is worth noting that Algorithm \ref{alg.AutoSC} can be easily implemented parallelly, which will reduce the time complexity to $O(\max(m,r)n^2+kmn)$. On the contrary, SSC, LRR, and their variants require iterative optimization and hence their time complexity is about $O(tmn^2)$, where $t$ denotes the iteration number and is often larger than $100$.

\renewcommand{\algorithmicrequire}{\textbf{Input:}}
\renewcommand{\algorithmicensure}{\textbf{Output:}}
\begin{algorithm}[h!]
\caption{AutoSC-GD with Only LSR and KLSR}
\label{alg.AutoSC}
\begin{algorithmic}[1]
\REQUIRE
$\bm{X}$, $k$, $\mathcal{F}$, $\Lambda$, $\mathcal{T}$
\STATE Normalize the columns of $\bm{X}$ to have unit $\ell_2$ norm.
\FOR{$f_u$ in $\mathcal{F}$}
	\FOR{$\lambda_i$ in $\Lambda$}
					\STATE Construct $\bm{C}$ by \eqref{eq.LSR_1} or \eqref{eq.KLR_1}.
			\FOR{$\tau_j$ in $\mathcal{T}$}
				\STATE $\bm{C}\leftarrow\vert \bm{C}\odot(\bm{1}-\bm{I})\vert$.
				\STATE Truncate $\bm{C}$ with parameter $\tau_j$.
				\STATE For $j=1,\ldots,n$, let $\bm{c}_j\leftarrow\bm{c}_j/\vert\bm{c}_j\vert_1$.
				\STATE $\bm{A}=(\bm{C}+\bm{C}^\top)/2$.
				\STATE $ \bm{L}=\bm{I}-\bm{D}^{-1/2}\bm{A}\bm{D}^{-1/2}$.
				\STATE Compute $\sigma_1,\ldots,\sigma_{k+1}$ and $\bm{v}_1,\ldots,\bm{v}_{k}$.
				\STATE $\Delta_{uij}=\textup{REG}(\bm{L})$, $\mathcal{V}_{uij}=[\bm{v}_1,\ldots,\bm{v}_{k}]$.
				
			\ENDFOR
	\ENDFOR
\ENDFOR
\STATE $\bm{Z}=\mathcal{V}_{\bar{u}\bar{i}\bar{j}}^\top$, where $\lbrace \bar{u},\bar{i},\bar{j}\rbrace=\textup{argmax}_{u,i,j}\Delta_{uij}$.
\STATE Normalize the columns of $\bm{Z}$ to have unit $\ell_2$ norm.
\STATE Perform k-means on $\bm{Z}$.
\ENSURE $k$ clusters: $C_1,\ldots,C_k$.
\end{algorithmic}
\end{algorithm}

\section{The algorithm of AutoSC+NSE}\label{sec_large}
See Algorithm \ref{alg.NSE}.

\begin{algorithm}[h!]
\caption{AutoSC+NSE}
\label{alg.NSE}
\begin{algorithmic}[1]
\REQUIRE
$\bm{X}$, $k$, $\mathcal{F}$, $\Theta$, $\hat{n}$.
\STATE Select $\hat{n}$ landmarks from $\bm{X}$ by k-means to form $\hat{\bm{X}}$. 
\STATE Apply AutoSC-GD or AutoSC-BO to  $\hat{\bm{X}}$ with $\mathcal{F}$ and $\Theta$.
\item  Get $\hat{\bm{Z}}$ from the best Laplacian matrix given by AutoSC-G or AutoSC-BO.
\STATE Use mini-batch Adam to solve \eqref{eq.NSE}.
\STATE Compute $\bm{Z}$ by \eqref{eq.NSE_Z}.
\STATE Perform k-means on $\bm{Z}$.
\ENSURE $k$ clusters: $C_1,\ldots,C_k$.
\end{algorithmic}
\end{algorithm}

\section{More theoretical results}\label{more_theory}

\subsection{Theoretical guarantee for KLSR}
\begin{definition}[Polynomial Deterministic Model]\label{def_kernel_reg}
The columns of $\bm{X}_0\in\mathbb{R}^{m\times n}$ are drawn from a union of $k$ different polynomials $\{g_j:\mathbb{R}^r\rightarrow\mathbb{R}^m, ~r<m\}_{j=1}^k$ of order at most $p$ and  are further corrupted by noise, say $\bm{X}=\bm{X}_0+\bm{E}$.  Denote the eigenvalue decomposition of the kernel matrix $\bm{K}$ of $\bm{X}$ as $\bm{K}=\bm{V}\bm{\Sigma}\bm{V}^\top$,  where $\bm{\Sigma}=\textup{diag}(\sigma_1,\ldots,\sigma_n)$ and $\sigma_1\geq\sigma_2\geq\cdots\sigma_n$.   Let $\gamma=\sigma_{d+1}/\sigma_d$. Denote $\bm{v}_i=(v_{i1},\ldots,v_{in})$ the $i$-th row of $\bm{V}$ and let $\bar{\bm{v}}_i=(v_{i1},\ldots,v_{id})$, where $d<n$. Suppose the following conditions hold: 1) for every $i\in[n]$,  the $\bar{\tau}$-th largest element of $\{\vert\bar{\bm{v}}_i^\top\bar{\bm{v}}_j\vert:j\in C_{\pi(i)}\}$ is greater than $\alpha$; 2) $\max_{i\in[n]}\max_{j\in [n]\setminus C_{\pi(i)}}\vert\bar{\bm{v}}_i^\top\bar{\bm{v}}_j\vert\leq\beta$; 3) $\max_{i,j,l}\vert v_{il}v_{jl}\vert\leq \mu$.
\end{definition}
Here we consider polynomials because they are easy to analyze and can well approximate smooth functions provided that $p$ is sufficiently large.  Clustering the columns of $\bm{X}$ given by Definition \ref{def_kernel_reg} according to the polynomials is actually a manifold clustering problem beyond the setting of subspace clustering.  Similar to the subspace detection property, we define
\begin{definition}[Manifold Detection Property]\label{def_manifold_dec}
A symmetric affinity matrix $\bm{A}$ obtained from $\bm{X}$ has manifold detection property if for all $i$,  the nonzero elements of $\bm{a}_i$ correspond to the columns of $\bm{X}$ lying on the same manifold as $\bm{x}_i$.
\end{definition}

The following theorem verifies the effectiveness of \eqref{eq.KLR_1} followed by the truncation operation in manifold detection.
\begin{theorem}\label{theorem_kernel_poly}
Suppose $\bm{X}$ and $\bm{K}$ are given by Definition \ref{def_kernel_reg} and $\bm{C}$ is given by \eqref{eq.KLR_1}, where the kernel function is a polynomial kernel of order $q$, $\textup{rank}(\bm{K}_0)=d$ ($\bm{K}_0$ is from $\bm{X}_0$), and
\begin{equation}
\tfrac{\left(\rho-\sqrt{\rho^2-4(2\mu d-\Delta)(2\mu n-2\mu d-\Delta)}\right)\sigma_d^2}{4\mu d-2\Delta}<\lambda<\tfrac{\left(\rho+\sqrt{\rho^2-4(2\mu d-\Delta)(2\mu n-2\mu d-\Delta)}\right)\sigma_d^2}{4\mu d-2\Delta}
\end{equation}

where $\rho=2\mu n\gamma^2-\Delta(1+\gamma^2)$.  
Then $d\leq k\binom{r+pq}{pq}$ and the $\bm{C}$ truncated by $\tau\leq \bar{\tau}$ has the manifold detection property.
\end{theorem}

In the theorem,  $\sigma_d$ can be much larger than $\sigma_{d+1}$ provided that the noise is small enough.  Then we get a wide range for $\lambda$.
Compared to Theorem \ref{theorem_ssd}, Theorem \ref{theorem_kernel_poly} allows a much larger $d$, which means the kernel method is able to handle more difficult clustering problems than the linear method.

\subsection{Theoretical analysis for NSE}\label{sec_prop_nse}

The following proposition shows that a small number of hidden nodes in NSE are sufficient to make the clustering succeed.
\begin{proposition}\label{prop_nn}
Suppose the columns (with unit $\ell_2$ norm) of $\bm{X}$ are drawn from a union of $k$ independent subspaces of dimension $r$: $\sum_{j=1}^k\textup{dim}(\mathcal{S}_j)=\textup{dim}(\mathcal{S}_1\cup \cdots \cup\mathcal{S}_k)=kr$. For $j=1,\ldots,k$, let $\bm{U}^{j}$ be the bases of $\mathcal{S}_j$ and $\bm{x}_i=\bm{U}^{j}\bm{v}_i$, if $\bm{x}_i\in\mathcal{S}_j$. Suppose $\max\lbrace \Vert {\bm{U}_{:l}^{i}}^\top\bm{U}^{j}\Vert: 1\leq l\leq r,1\leq i\neq j\leq k\rbrace\leq\mu$. Suppose that for all $i=1,\ldots n$, $\max \lbrace v_{1i},\ldots,v_{ri}\rbrace>\mu$.
Then there exist $\bm{W}_1\in\mathbb{R}^{d\times m}$, $\bm{W}_2\in\mathbb{R}^{k\times d}$, $\bm{b}_1\in\mathbb{R}^{d}$, and $\bm{b}_2\in\mathbb{R}^{k}$ such that performing k-means on $\bm{Z}$ given by \eqref{eq.NSE_Z} identifies the clusters correctly, where $d=kr$.
\end{proposition}

\section{More about the experiments}
\subsection{Dataset description}\label{sec_datasets}
The description for the benchmark image datasets considered in this paper are as follows.
\begin{itemize}
\item \textbf{Extended Yale B Face} \citep{Dataset_ExtendYaleB} (Yale B for short): face images (192$\times$168) of 38 subjects. Each subject has about 64 images under various illumination conditions. We resize the images into $32\times 32$.
\item \textbf{ORL Face} \citep{ORL_face}: face images (112$\times$92)
of 40 subjects. Each subject has 10 images with different poses and facial expressions.  We resize the images into 32$\times$32.
\item \textbf{COIL20} \citep{Dataset_COIL20}: images ($32\times 32$) of 20 objects. Each object has 72 images of different poses.
\item \textbf{AR Face} \citep{ARfacedata}: face images (165$\times$120) of 50 males and 50 females. Each subject has 26 images with different facial expressions, illumination conditions, and occlusions. We resize the images into $42\times 30$.
\item \textbf{MNIST} \citep{lecun1998gradient}: 70,000 grey images ($28\times 28$) of handwritten digits $0-9$.
\item \textbf{MNIST-1k(10k)}: a subset of  MNIST containing 1000(10000) samples, 100(1000) randomly selected samples per class.
\item  \textbf{Fashion-MNIST} \citep{xiao2017fmnist}: 70,000 gray images ($28\times 28$) of 10 types of fashion product. 
\item \textbf{Fashion-MNIST-1k(10k)}: a subset of  Fashion-MNIST containing 1000(10000) samples, 100(1000) randomly selected samples per class.
\item \textbf{MNIST-feature}: following the same procedures of  \citep{chen2020stochastic}, we compute a feature vector of dimension 3,472 using the scattering convolution network \citep{bruna2013invariant} and then reduce the dimension to 500 using PCA. 
\item \textbf{Fashion-MNIST-feature}: similar to  MNIST-feature.
\item \textbf{GTSRB} \citep{stallkamp2012man}: consisting of 12,390 images of street signs in 14 categories.  Following \citep{chen2020stochastic}, we extract a 1568-dimensional HOG feature, and reduce the dimension to 500 by PCA. 
\end{itemize}

All experiments are conducted in MATLAB on a MacBook Pro with 2.3 GHz Intel i5 Core and 8GB RAM.

\subsection{Hyperparameter settings for the small datasets}\label{app_hyper_small}
We select $\lambda$  from $\lbrace 0.01, 0.02, 0.05,0.1,0.2,\ldots,0.5 \rbrace$ for SSC, LRR, and KSSC. The $\lambda$ in BDR is chosen from $\lbrace 5,10,20,\ldots,80\rbrace$. The $\gamma$ in BDR-B and BDR-Z is chosen from $\lbrace 0.01,0.1,1\rbrace$. The parameter $s$ in SSC-OMP is chosen from $\lbrace 3,4,\ldots,15\rbrace$. We report the results of these methods with their best hyperparameters.
In AutoSC, we set $\Lambda=\lbrace 0.01, 0.1, 1 \rbrace$ and $\mathcal{T}=\lbrace5,6,\ldots,15\rbrace$. In AutoSC-BO,  we consider two models: 1) Gaussian kernel similarity; 2) KLSR with polynomial kernel; 3)  KLSR with Gaussian kernel,  in which the hyperparameters of kernels are optimized adaptively. Then we needn't to consider LSR explicitly because it is a special case of KLSR with polynomial kernel. See Appendix \ref{appendix_ssc}.

\subsection{Clustering results in terms of  NMI}\label{sec_nmi}
In addition to the clustering accuracy reported in Table \ref{tab_five}, here we also compare the normalized mutual information (NMI) in Table \ref{tab_five_nmi}.  We see that the comparative performance of all methods are similar to the results in Table \ref{tab_five} and our methods AutoSC-GD and AutoSC-BO outperformed other methods in almost all cases. 
\begin{table*}[h]
\centering
\caption{Normalized Mutual Information on the six small datasets}\label{tab_five_nmi}
\begin{small}
\begin{tabular}{c|ccccccccc}\hline
 & SSC & LRR & EDSC & KSSC & SSC-OMP&BDR-Z	&BDR-B	&AutoSC-GD&AutoSC-BO\\ \hline
Yale B& 0.817 &	0.703	& 0.835 &0.730	&0.841	&0.666	&0.743	&\textbf{0.919}& \textbf{0.928}\\ \hline
ORL&0.849 	&0.872&	0.856 &0.872&0.815		&0.875	 & 0.865& \textbf{0.907}   &\textbf{0.903}    \\ \hline
COIL20&0.954&	0.706&0.843	&\textbf{0.983}&	0.671	&0.843&	0.873&	0.897&\textbf{0.963}   \\ \hline
AR&0.818&	0.872& 0.825	&0.809&	0.691&	0.865&	0.861&	{\textbf{0.887}}& \textbf{0.904}\\ \hline
MNIST-1k&0.612	&0.538	& 0.631&0.626	&0.546&	0.634	&0.580	&\textbf{0.667}& \makecell{\textbf{0.652}}\\
\hline
Fashion-MNIST-1k&0.616&0.601&0.621&0.621&0.559&0.614&0.605&\textbf{0.633}& \textbf{0.629}\\ \hline
\end{tabular}
\end{small}
\end{table*}

\subsection{The stability of AutoSC}\label{appendix_stab}
\begin{figure}[h]
\centering
\includegraphics[width=13cm]{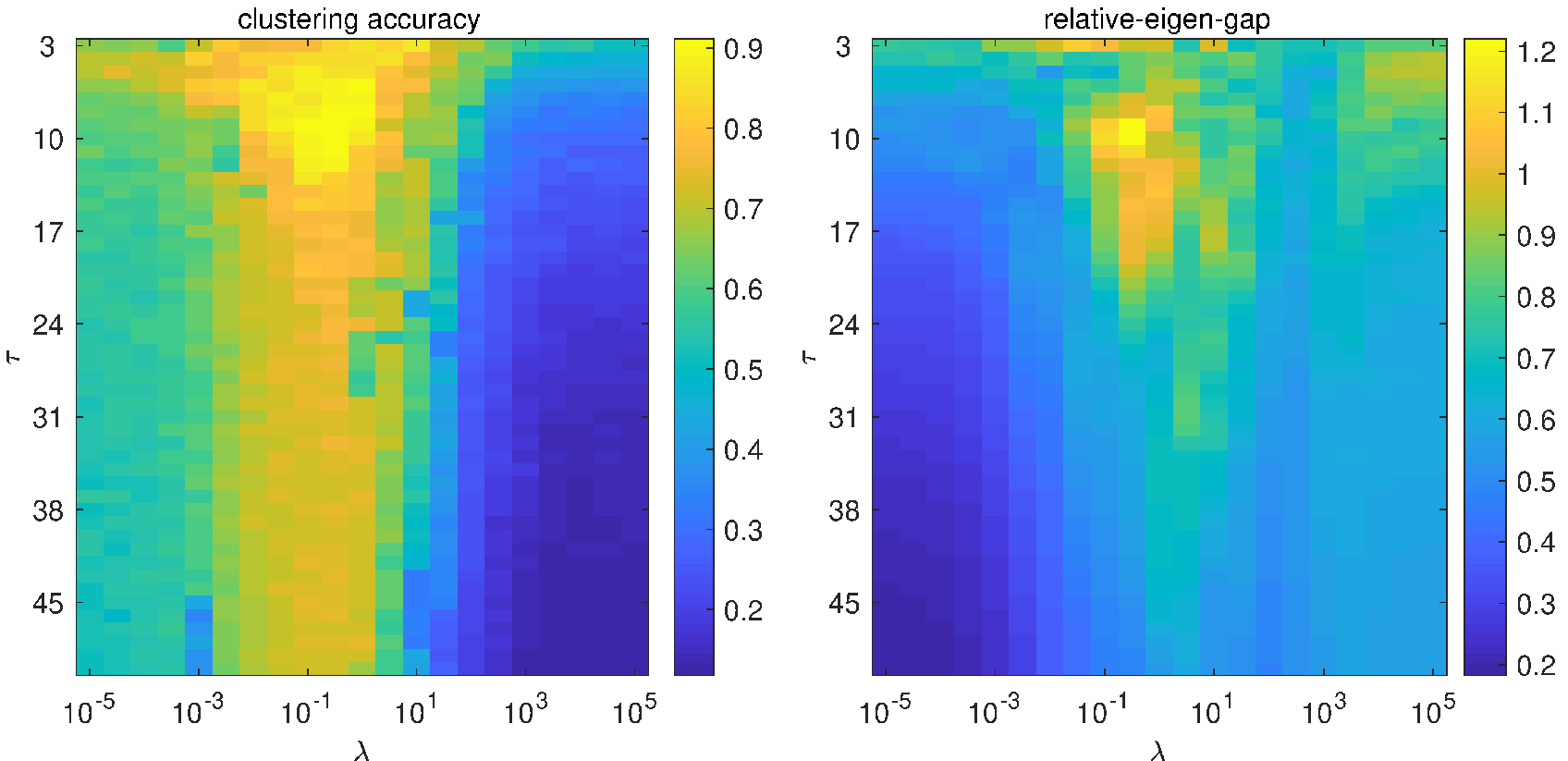}
\caption{Visualization of the clustering accuracy and the corresponding relative-eigen-gap when a large search space is used.}\label{fig_par2d}
\end{figure}

Though we have used a relatively compact search space in AutoSC to reduce the highly unnecessary computational cost, the search space can be arbitrarily large.  Figure \ref{fig_par2d} shows the clustering accuracy and the corresponding relative-eigen-gap. We can see that the region with highest relative-eigen-gap is in accordance with the region with highest clustering accuracy.

\subsection{More about AutoSC-BO in the experiments}\label{appendix_ssc}
For SSC, we consider the following problem
\begin{equation}\label{eq.KSSC}
\mathop{\textup{minimize}}_{\bm{C}}\ \tfrac{1}{2}\textup{Tr}\left(\bm{K}-2\bm{K}\bm{C}+\bm{C}^\top\bm{K}\bm{C}\right)+{\lambda}\Vert\bm{C}\Vert_1, 
\end{equation}
where $\bm{K}$ is an $n\times n$ kernel matrix with $[\bm{K}]_{ij}=k(\bm{x}_i,\bm{x}_j)$.  Note that when we use a linear kernel function,  \eqref{eq.KSSC} reduces to the vanilla SSC. We solve the optimization via alternating direction method of multipliers (ADMM) \citep{ADMM}, where the Lagrange parameter is 0.1 and the maximum number of iterations is 500. In this study, we consider polynomial kernel and Gaussian kernel, and optimize all hyperparameters including the order of the polynomial kernel.  Particularly,  for Gaussian kernel, we set $\varsigma=\frac{\xi}{n^2}\sum_{ij}\Vert\bm{x}_i-\bm{x}_j\Vert$ and optimize $\xi$. The search space for the hyperparameters are as follows: $10^{-3}\leq\lambda\leq 1$, $5\leq\tau\leq 50$, $0\leq b\leq 10^3$, $1\leq q\leq 5$,  $0.5\leq\xi\leq 5$.

In addition to Figure 2 of the main paper, here we report the best hyperparameters of the four models found by AutoSC-BO in Table \ref{tab_bestpar}. It can be found that the accuracy of KLSR with a linear kernel is higher than other models, which is consistent with its highest reg.  

\renewcommand\tabcolsep{10pt}
\begin{table}[h!]
\centering
\caption{The best hyperparameters and the corresponding clustering accuracy given by AutoSC$_{BO}$ on the first 10 subjects of YaleB Face dataset.}\label{tab_bestpar}
\begin{tabular}{l|c|c|c}
\hline
method&hyperparameters&reg&accuracy\\
\hline
\makecell{KLSR\\(Polynomial)}&\makecell{$\lambda=0.207$, $b=19.09$,\\$q=1$, $\tau=5$}&2.379&0.966\\ \hline
\makecell{KLSR\\(Gaussian)}&\makecell{$\lambda=0.013$,\\$\xi=4.92$, $\tau=5$}&2.217&0.963\\ \hline
\makecell{KSSC\\(Polynomial)}&\makecell{$\lambda=0.519$, $b=44.57$,\\$q=2$, $\tau=5$}&1.388&0.859\\ \hline
\makecell{KSSC\\(Gaussian)}&\makecell{$\lambda=0.0011$, \\$\xi=4.97$, $\tau=6$}&0.892&0.584\\
\hline
\end{tabular}
\end{table}

\subsection{Hyperparameter settings of large-scale clustering}\label{sec_hyper_large}
On MNIST-10k, MNIST, Fashion-MNIST-10k, and Fashion-MNIST, the parameter settings of \citep{chen2011large}, SSSC \citep{peng2013scalable}, SSC-OMP \citep{you2016scalable}, and S$^5$C \citep{matsushima2019selective}, and S$^3$COMP-C \citep{chen2020stochastic}, and AutoSC+NSE are shown in Table \ref{tabpara}. These hyper parameters  have been determined via grid search and the best (as possible) values are used.

\begin{table}[h!]
\centering
\caption{Hyper-parameter settings of the compared methods on MNIST-10k, MNIST, Fashion-MNIST-10k, and Fashion-MNIST. $s$ denotes the number of landmark data points. In the optimization (mini-batch Adam) of AutoSC+NSE, the epoch number, batch size, and step size are 200, 128,  and $10^{-3}$ respectively.}\label{tabpara}
\begin{tabular}{l|l}\hline
LSC-K& $s=1000$, $r=3$\\ \hline
SSSC& $s=1000$, $\lambda=0.01$\\ \hline
SSC-OMP	& $K=10$ (sparsity) \\ \hline
S$^5$C & $s=1000$, $\lambda=0.1$ or $0.2$\\ \hline
S$^3$COMP-C & $T=20$, $\lambda=0.4$, $\delta=0.9$\\ \hline
AutoSC+NSE & $s=1000$, $d=200$, $\gamma=10^{-5}$\\ \hline
AutoSC$_{BO}$+NSE& $s=1000$, $d=200$, $\gamma=10^{-5}$\\ \hline
\end{tabular}
\end{table}

\subsection{Influence of hyper-parameters in AutoSC+NSE}\label{sec_more_AutoSC_nse}
\begin{figure}[h!]
\centering
\includegraphics[width=8cm]{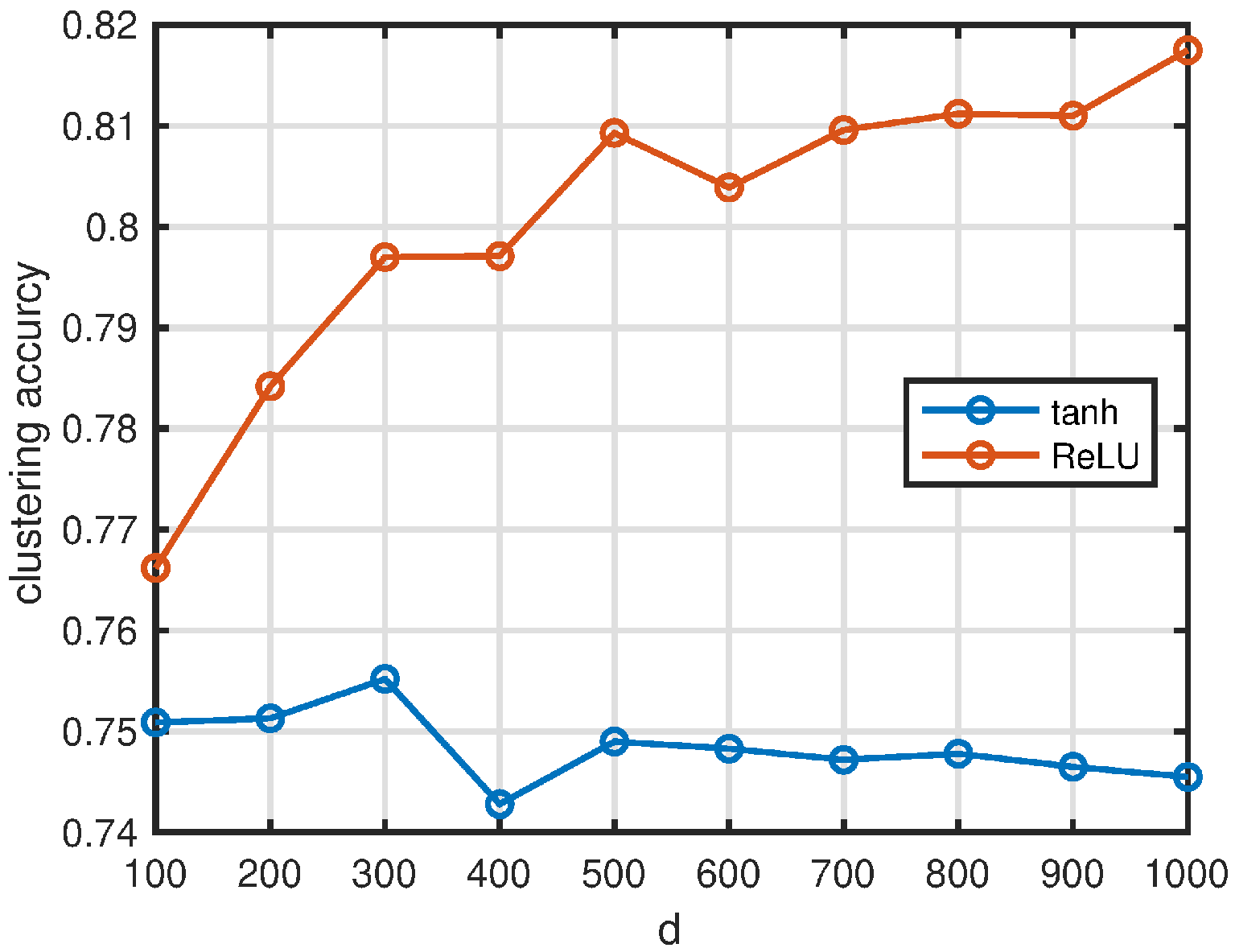}
\caption{ReLU v.s. tanh (hyperbolic tangent) in the hidden layer of AutoSC+NSE on MNIST.  When using ReLU, we set $\gamma=10^{-5}$ and $\alpha=10^{-3}$ (the step size in Adam). When using tanh, we set $\gamma=10^{-3}$ and $\alpha=10^{-2}$, which perform best in this case. Notice that the clustering accuracy when using ReLU is higher than 0.78 in almost all cases, which is higher than the value (say $0.755$) we reported in the main paper. The reason is that in the main paper, we reported the mean value of 10 repeated trials but here we report the value of a single trial.}\label{fig.actfun}
\end{figure}

We investigate the effects of the type of activation function and the number ($d$) of nodes in the hidden layer of NSE. For convenience, we used a fixed random seed of MATLAB (rng(1)). Figure \ref{fig.actfun} shows the clustering accuracy on MNIST given by AutoSC+NSE with different activation function and different $d$. We see that ReLU outperformed tanh consistently. The reason is that the nonlinear mapping $g$ from the data space to the eigenspace of the Laplacian matrix is nonsmooth and ReLU is more effective than tanh in approximating nonsmooth functions. In addition, when $d$ increases, the clustering accuracy of AutoSC+NSE with ReLU often becomes higher because a wider network often has a higher ability of function approximation.

\begin{figure}[h!]
\centering
\includegraphics[width=8cm]{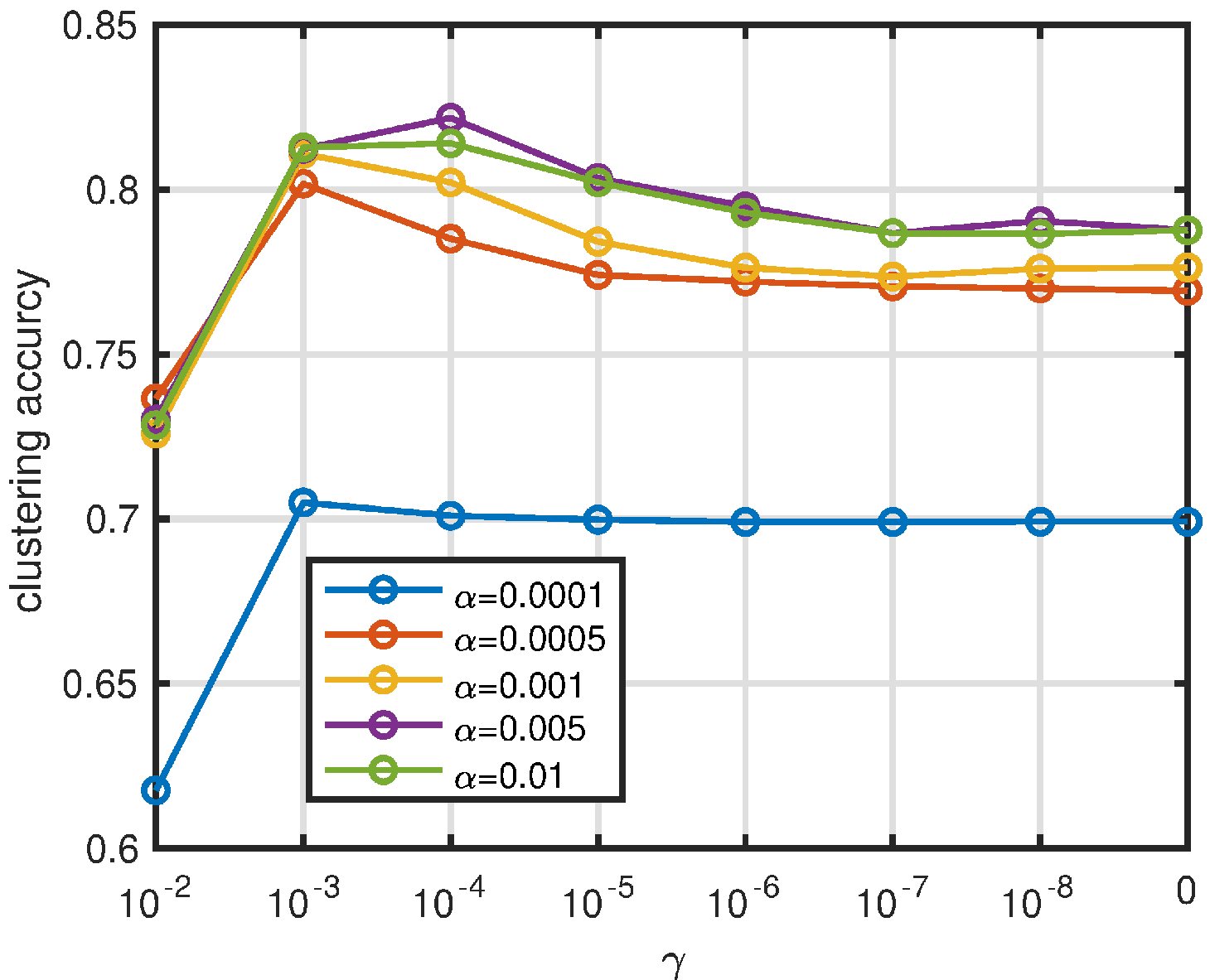}
\caption{Influence of $\gamma$ and $\alpha$ in AutoSC+NSE on MNIST. We set $d=200$ and use ReLU.}
\label{fig_gamma}
\end{figure}

Figure \ref{fig_gamma} shows the clustering accuracy on MNIST given by AutoSC+NSE with different $\gamma$ and $\alpha$. When $\alpha$ is too small (say $10^{-4}$, the clustering accuracy is low, because the training error is quite large in 200 epochs. In fact, by  increasing the training epochs, the clustering accuracy can be improved, which however will increase the time cost. When $\alpha$ is relatively large, the clustering accuracy is often higher than 0.755.
On the other hand, AutoSC+NSE is not sensitive to $\gamma$ provided that it is not too large.

\begin{figure}[h!]
\centering
\includegraphics[width=8cm]{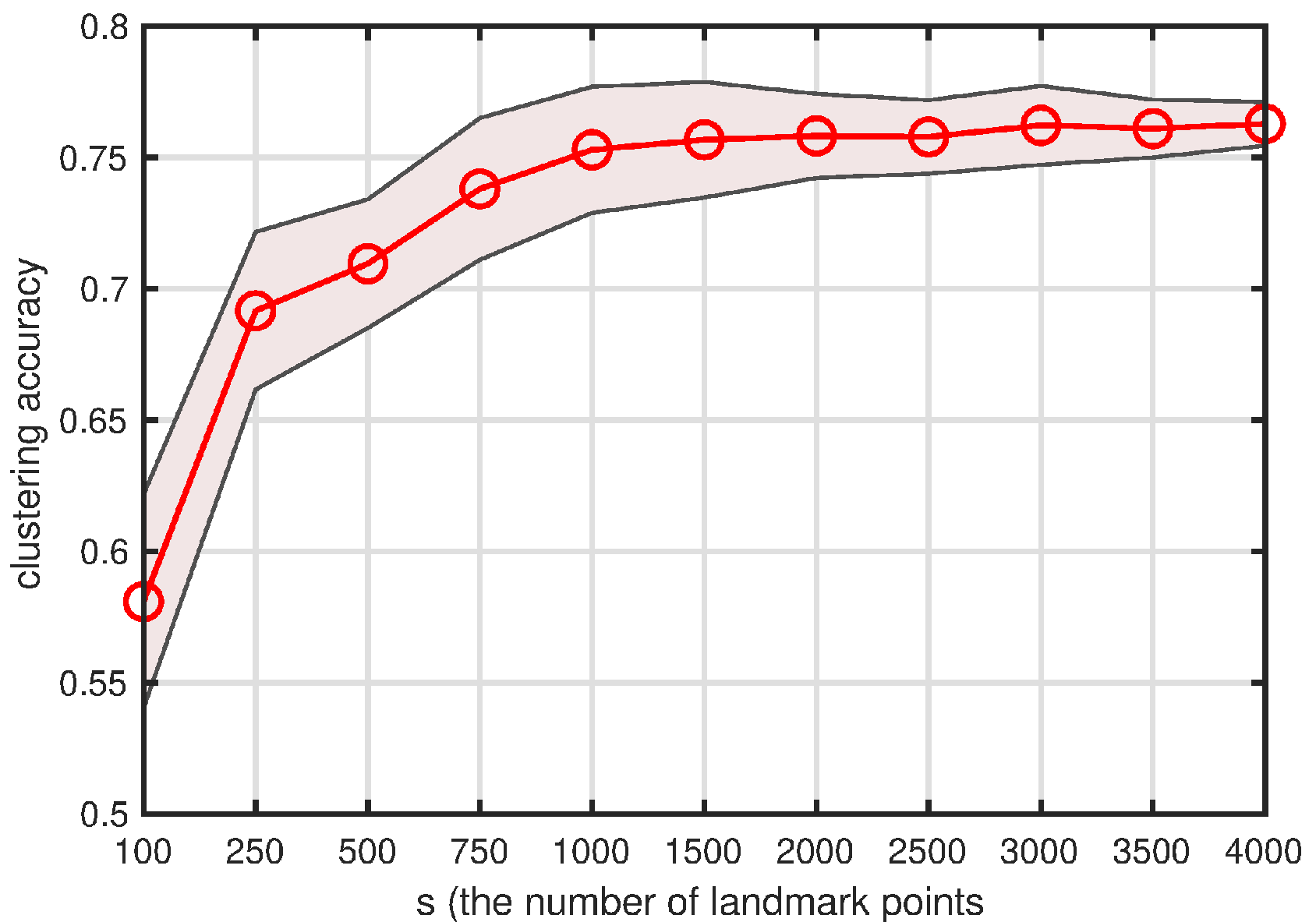}
\caption{Influence of the number of landmark points in AutoSC+NSE on MNIST. We set $d=200$, $\gamma=10^{-5}$, and $\alpha=10^{-3}$. The shadow denotes the standard deviation of 10 trials.}
\label{fig_s}
\end{figure}

Figure \ref{fig_s} shows the mean value and standard deviation (10 repeated trials) of the clustering accuracy on MNIST given by AutoSC+NSE with different number (denoted by $s$) of landmark points. It can be found that when the $s$ increases, the clustering accuracy increases and its standard deviation becomes smaller. When $s$ is large enough, the improvement is not significant.

\section{Proof for the theoretical results}
\subsection{Proof for Claim \ref{claim_sumk}}
\begin{proof}
The stochastic transition matrix of $G$ is defined as
\begin{equation}
\bm{P}=\bm{D}^{-1}\bm{A}.
\end{equation}
In \citep{meila2001multicut}, it was showed that
\begin{equation}\label{eq.MNCut01}
\textup{MNCut}(\mathcal{C})\geq k-\sum_{i=1}^k\varrho_i(\bm{P}),
\end{equation}
where $\varrho_i(\bm{P})$ denotes the $i$-th largest eigenvalue of $\bm{P}$ and $1=\varrho_1(\bm{P})\geq \varrho_2(\bm{P})\geq\cdots\varrho_k(\bm{P})$.
According to Lemma 3 of \citep{meila2001multicut}, we have
\begin{equation}\label{eq.PL}
\sigma_i(\bm{L})=1-\varrho_i(\bm{P}),\quad  \forall i=1,\ldots,n.
\end{equation}
Substituting \eqref{eq.PL} into \eqref{eq.MNCut01}, we have
\begin{equation}\label{eq.MNCu>>}
\textup{MNCut}(\mathcal{C})\geq \sum_{i=1}^k\sigma_i(\bm{L}).
\end{equation}
\end{proof}

\begin{remark}
$\mathcal{C}$ can be any partition of the nodes of $G$. Let $\mathcal{C}^\ast$ be the optimal partition. Then $\textup{MNCut}(\mathcal{C}^\ast)=\sum_{i=1}^k\sigma_i(\bm{L})$. If $\sum_{i=1}^k\sigma_i(\bm{L})=0$, there are no connections (edges) among $C_1^\ast, \ldots, C_k^\ast$.
\end{remark}

\subsection{Proof for Claim \ref{claim_k+1}}

\begin{proof}
For $i=1,\ldots,k$, we aim to partition $C_i$ into two subsets, denoted by $C_i^1$ and $C_i^2$. Then we define
\begin{equation}
\textup{MNCut}(C_i)=\dfrac{Cut(C_i^1,C_i^2)}{Vol(C_i^1)}+\dfrac{Cut(C_i^2,C_i^1)}{Vol(C_i^2)}.
\end{equation}
It follows that
\begin{equation}
\textup{MNCut}(C_i)\geq\sum_{j=1}^2\sigma_j(\bm{L}_{C_i})\geq\sigma_2(\bm{L}_{C_i})=ac(C_i),
\end{equation}
where $\bm{L}_{C_i}$ denotes the Laplacian matrix of $C_i$ an $ i=1,\ldots,k$.
Since $\sigma_{k+1}(\bm{L})=\min\lbrace ac(C_1),\ldots,ac(C_k) \rbrace$, we have
\begin{equation}
\min_{1\leq i\leq k}\textup{MNCut}(C_i)\geq \sigma_{k+1}(\bm{L}).
\end{equation}
Therefore, $\sigma_{k+1}(\bm{L})$ measures the least connectivity of $C_1,\ldots,C_k$. This finished the proof.
\end{proof}

\begin{remark}
When $\sigma_{k+1}(\bm{L})$ is large, the connectivity in each of $C_1,\ldots,C_k$ is strong. Otherwise,  the connectivity in each of $C_1,\ldots,C_k$ is weak. When $\sigma_{k+1}(\bm{L})=0$, at least one of $C_1,\ldots,C_k$ contains at least two components, which means the nodes of $G$ can be partitioned into $k+1$ or more clusters.
\end{remark}

\subsection{Proof for Theorem \ref{the_1}}

\begin{proof}
According to Theorem 1 of \citep{regsc2005}, we have
\begin{equation}\label{eq.Meila_the1}
\textup{dist}(\mathcal{C}, \mathcal{C}')<\dfrac{3\delta}{\sigma_{k+1}(\bm{L})-\sigma_k(\bm{L})}.
\end{equation}
Since $\textup{reg}(\bm{L})=\dfrac{\sigma_{k+1}(\bm{L})-\frac{1}{k}\sum_{i=1}^k\sigma_i(\bm{L})}{\frac{1}{k}\sum_{i=1}^k\sigma_i(\bm{L})+\epsilon}$, we have
\begin{equation}\label{eq.Lgap}
\sigma_{k+1}(\bm{L})-\sigma_k(\bm{L})=\textup{reg}(\bm{L})(\bar{\sigma}+\epsilon)+\bar{\sigma}-\sigma_k(\bm{L}),
\end{equation}
where $\bar{\sigma}=\frac{1}{k}\sum_{i=1}^k\sigma_i(\bm{L})\geq\epsilon$.
Invoking \eqref{eq.Lgap} into \eqref{eq.Meila_the1}, we arrive at
\begin{equation*}
\begin{aligned}
\textup{dist}(\mathcal{C}, \mathcal{C}')&<\dfrac{3\delta}{\textup{reg}(\bm{L})(\bar{\sigma}+\epsilon)+\bar{\sigma}-\sigma_k(\bm{L})}\\
&\leq \dfrac{3\delta}{2\epsilon\textup{reg}(\bm{L})+\bar{\sigma}-k\bar{\sigma}}\\
&\leq \dfrac{3\delta}{2\epsilon\textup{reg}(\bm{L})+(1-k)\eta\epsilon}\\
&\leq \dfrac{1.5\delta\epsilon^{-1}}{\textup{reg}(\bm{L})+(1-k)\eta/2}.
\end{aligned}
\end{equation*}
This finished the proof.
\end{proof}

\subsection{Proof for Proposition \ref{prop_kernel}}

\begin{proof}
Since $\hat{\bm{c}}$ is the optimal solution, we have
$$\phi(\bm{x}_i)^\top\left(\phi(\bm{y})-\phi(\bm{X})\hat{\bm{c}}\right)+\lambda\hat{c}_i=0,$$
$$\phi(\bm{x}_j)^\top\left(\phi(\bm{y})-\phi(\bm{X})\hat{\bm{c}}\right)+\lambda\hat{c}_j=0.$$
It follows that
\begin{equation}
\begin{aligned}
\Vert\hat{c}_i-\hat{c}_j\Vert&=\Vert \left(\phi(\bm{x}_i)-\phi(\bm{x}_j)\right)^\top\left(\phi(\bm{y})-\phi(\bm{X})\hat{\bm{c}}\right)\Vert\\
&\leq \Vert \phi(\bm{x}_i)-\phi(\bm{x}_j)\Vert \Vert\phi(\bm{y})-\phi(\bm{X})\hat{\bm{c}}\Vert\\
&= \sqrt{k(\bm{x}_i,\bm{x}_i)-2k(\bm{x}_i,\bm{x}_j)+k(\bm{x}_j,\bm{x}_j)}\\
&\quad \times \Vert\phi(\bm{y})-\phi(\bm{X})\hat{\bm{c}}\Vert\\
&=\sqrt{2-2k(\bm{x}_i,\bm{x}_j)}\Vert\phi(\bm{y})-\phi(\bm{X})\hat{\bm{c}}\Vert\\
&\leq \sqrt{2-2k(\bm{x}_i,\bm{x}_j)}\Vert\phi(\bm{y})\Vert\\
&=\sqrt{2-2\exp\left(-\frac{\Vert\bm{x}_i-\bm{x}_j\Vert^2}{2\varsigma^2}\right)}.\\
\end{aligned}
\end{equation}
In the second and last equalities, we used the fact that $\Vert\phi(\bm{y})\Vert=\Vert\phi(\bm{x})\Vert=1$. In the second inequality, we used the fact that
$ \dfrac{1}{2}\Vert\phi(\bm{y})-\phi(\bm{X})\hat{\bm{c}}\Vert^2+\dfrac{\lambda}{2}\Vert\hat{\bm{c}}\Vert^2\leq  \dfrac{1}{2}\Vert\phi(\bm{y})-\phi(\bm{X})\bm{0}\Vert^2+\dfrac{\lambda}{2}\Vert\bm{0}\Vert^2=\dfrac{1}{2}\Vert\phi(\bm{y})\Vert^2$ because $\hat{\bm{c}}$ is the optimal solution.
\end{proof}

\subsection{Proof for Theorem \ref{theorem_ssd}}
\begin{proof}
Invoking the SVD of $\bm{X}$ into the closed-form solution of LSR, we get 
\begin{equation}
\bm{C}=\bm{V}\textup{diag}\left(\frac{\sigma_1^2}{\sigma_1^2+\lambda},\ldots,\frac{\sigma_n^2}{\sigma_n^2+\lambda}\right)\bm{V}^\top.
\end{equation}
It means
\begin{equation}
\begin{aligned}
c_{it}=&\sum_{l=1}^n\frac{v_{il}v_{jl}\sigma_l^2}{\sigma_l^2+\lambda}\\
=&\bar{\bm{v}}_i^\top\bar{\bm{v}}_t-\sum_{l={1}}^d\frac{v_{il}v_{tl}\lambda}{\sigma_l^2+\lambda}
+\sum_{l={d+1}}^n\frac{v_{il}v_{tl}\sigma_l^2}{\sigma_l^2+\lambda}.
\end{aligned}
\end{equation}
Suppose $j\in C_{\pi(i)}$ and $k\in [n]\setminus C_{\pi(i)}$. We have
\begin{equation}
\begin{aligned}
&\vert c_{ij}\vert-\vert c_{ik}\vert\\
=&\left\vert \bar{\bm{v}}_i^\top\bar{\bm{v}}_j-\sum_{l={1}}^d\frac{v_{il}v_{jl}\lambda}{\sigma_l^2+\lambda}
+\sum_{l={d+1}}^n\frac{v_{il}v_{jl}\sigma_l^2}{\sigma_l^2+\lambda}\right\vert\\
&-\left\vert \bar{\bm{v}}_i^\top\bar{\bm{v}}_k-\sum_{l={1}}^d\frac{v_{il}v_{kl}\lambda}{\sigma_l^2+\lambda}
+\sum_{l={d+1}}^n\frac{v_{il}v_{kl}\sigma_l^2}{\sigma_l^2+\lambda}\right\vert\\
\geq &\left\vert \bar{\bm{v}}_i^\top\bar{\bm{v}}_j\right\vert-\left\vert \bar{\bm{v}}_i^\top\bar{\bm{v}}_k\right\vert-\left\vert \sum_{l={1}}^d\frac{v_{il}v_{jl}\lambda}{\sigma_l^2+\lambda} \right\vert-\left\vert \sum_{l={1}}^d\frac{v_{il}v_{kl}\lambda}{\sigma_l^2+\lambda}
\right\vert\\
&-\left\vert \sum_{l={d+1}}^n\frac{v_{il}v_{jl}\sigma_l^2}{\sigma_l^2+\lambda}\right\vert-\left\vert\sum_{l={d+1}}^n\frac{v_{il}v_{kl}\sigma_l^2}{\sigma_l^2+\lambda}\right\vert\\
\geq&\left\vert \bar{\bm{v}}_i^\top\bar{\bm{v}}_j\right\vert-\left\vert \bar{\bm{v}}_i^\top\bar{\bm{v}}_k\right\vert-
2\mu\sum_{l={1}}^d\frac{\lambda}{\sigma_l^2+\lambda}-2\mu \sum_{l={d+1}}^n\frac{\sigma_l^2}{\sigma_l^2+\lambda}\\
\geq &\left\vert \bar{\bm{v}}_i^\top\bar{\bm{v}}_j\right\vert-\beta-
\frac{2\mu d\lambda}{\sigma_d^2+\lambda}-\frac{2\mu a\sigma_{d+1}^2}{\sigma_{d+1}^2+\lambda},
\end{aligned}
\end{equation}
where $a=\min(m,n)-d=m-d$. 

To ensure that there exist at least $\bar{\tau}$ elements of $\{\vert c_{ij}\vert: j\in C_{\pi(i)}\}$ greater than $\vert c_{ik}\vert$ for all $k\in [n]\backslash C_{\pi(i)}$, we need
\begin{equation}
\left\vert \bar{\bm{v}}_i^\top\bar{\bm{v}}_j\right\vert-\beta-
\frac{2\mu d\lambda}{\sigma_d^2+\lambda}-\frac{2\mu a\sigma_{d+1}^2}{\sigma_{d+1}^2+\lambda}>0
\end{equation}
holds at least for $\bar{\tau}$ different $j$,  where $j\in C_{\pi(i)}$. It is equivalent to ensure that
\begin{equation}\label{eq_proof_ineq_0}
\alpha-\beta-
\frac{2\mu d\lambda}{\sigma_d^2+\lambda}-\frac{2\mu a\sigma_{d+1}^2}{\sigma_{d+1}^2+\lambda}>0.
\end{equation}
We rewrite \eqref{eq_proof_ineq_0} as
\begin{equation}\label{eq_proof_ineq_1}
u_1\lambda^2+u_2\lambda+u_3>0,
\end{equation}
where $u_1=\alpha-\beta-2\mu d$, $u_2=(\alpha-\beta)(\sigma_d^2+\sigma_{d+1}^2)-2\mu(d+a)\sigma_{d+1}^2$,  and $u_3=(\alpha-\beta-2\mu a)\sigma_d^2\sigma_{d+1}^2$.

The definition of $\mu$, $\alpha$, and $\beta$ imply $u_1<0$. Then we solve \eqref{eq_proof_ineq_1} and obtain
\begin{equation}
\left\lbrace
\begin{matrix}
\lambda>\frac{2\mu m\sigma_{d+1}^2-(\alpha-\beta)(\sigma_d^2+\sigma_{d+1}^2)-\sqrt{w}}{2(2\mu d-(\alpha-\beta))}\\
\lambda<\frac{2\mu m\sigma_{d+1}^2-(\alpha-\beta)(\sigma_d^2+\sigma_{d+1}^2)+\sqrt{w}}{2(2\mu d-(\alpha-\beta))}
\end{matrix}
\right.
\end{equation}
where $w=u_2^2-4u_1u_3$. To simplify the notations, we let $\Delta=\alpha-\beta$, $\sigma_{d+1}=\gamma\sigma_d$ and get
\begin{equation}
\left\lbrace
\begin{matrix}
\lambda>\frac{\left(2\mu m\gamma^2-\Delta(1+\gamma^2)-\sqrt{\left(\Delta(1+\gamma^2)-2\mu m\gamma^2\right)^2-4(\Delta-2\mu d)(\Delta-2\mu m+2\mu d)}\right)\sigma_d^2}{4\mu d-2\Delta}\\
\lambda<\frac{\left(2\mu m\gamma^2-\Delta(1+\gamma^2)+\sqrt{\left(\Delta(1+\gamma^2)-2\mu m\gamma^2\right)^2-4(\Delta-2\mu d)(\Delta-2\mu m+2\mu d)}\right)\sigma_d^2}{4\mu d-2\Delta}
\end{matrix}
\right.
\end{equation}
Further, let $\rho=2\mu m\gamma^2-\Delta(1+\gamma^2)$, we arrive at
\begin{equation}\label{eq_proof_lambda_last}
\left\lbrace
\begin{matrix}
\lambda>\frac{\left(\rho-\sqrt{\rho^2-4(\Delta-2\mu d)(\Delta-2\mu m+2\mu d)}\right)\sigma_d^2}{4\mu d-2\Delta}\\
\lambda<\frac{\left(\rho+\sqrt{\rho^2-4(\Delta-2\mu d)(\Delta-2\mu m+2\mu d)}\right)\sigma_d^2}{4\mu d-2\Delta}
\end{matrix}
\right.
\end{equation}
That means, if \eqref{eq_proof_lambda_last} holds,  for every $i$, the indices of the largest $\bar{\tau}$ absolute elements in the $i$-th column of $C$ are in $C_{\pi(i)}$. Therefore, the truncation operation with parameter $\tau\leq \bar{\tau}$ ensures the subspace detection property. 
This finished the proof.

\end{proof}

\subsection{Proof for Proposition \ref{prop_EGGS_G}}
\begin{proof}
The condition of reg means
\begin{equation*}
\dfrac{\sigma_{k+1}(\bm{L})-\frac{1}{k}\sum_{i=1}^k\sigma_i(\bm{L})}{\frac{1}{k}\sum_{i=1}^k\sigma_i(\bm{L})+\epsilon}=\dfrac{\sigma_{k+1}(\bm{L})}{\epsilon}>0.
\end{equation*}
For convenience, denote $\vartheta=\frac{1}{k}\sum_{i=1}^k\sigma_i(\bm{L})$. We have
\begin{equation*}
-\vartheta \epsilon=\vartheta\sigma_{k+1}.
\end{equation*}
It indicates $\vartheta=0$ and $\sigma_{k+1}\neq 0$.
Therefore the graph has exactly $k$ connected components. Since the subspace or manifold detection property hold for $\bm{A}$,  each component of $G$ is composed of the columns of $\bm{X}$ in the same subspace or manifold. Thus, all the columns of $\bm{X}$ in the same subspace or manifold must be in the same component.  Otherwise, the number of connected components is larger than $k$.
\end{proof}

\subsection{Proof for Theorem \ref{theorem_kernel_poly}}
The proof is nearly the same as that for Theorem \ref{theorem_ssd}, except that $d<n$ and $\textup{rank}(\bm{K}_0)\leq k\binom{r+pq}{pq}$, where $\bm{K}_0=\phi(\bm{X}_0)^\top\phi(\bm{X}_0)$. In this case,   $\bm{K}$ can be well approximately by a low-rank matrix of rank at most $k\binom{r+pq}{pq}$ provided that the noise is small enough. More details about $\bm{K}_0$ can be found in \citep{fan2020polynomial}.

\subsection{Proof for Proposition \ref{prop_nn}}

\begin{proof}
We only need to provide an example of $\bm{W}_1\in\mathbb{R}^{d\times m}$, $\bm{W}_2\in\mathbb{R}^{k\times d}$, $\bm{b}_1\in\mathbb{R}^{d}$, and $\bm{b}_2\in\mathbb{R}^{k}$, where $d=kr$, such that the clusters can be recognized by k-means.

We organize the rows of $\bm{W}_1$ into $k$ groups: $\bm{W}_1^j\in\mathbb{R}^{r\times m}$, $j=1,\ldots,k$. Let $\bm{W}_1^j={\bm{U}^{j}}^\top$, $j=1,\ldots,k$. Let $\bm{W}_1\bm{x}_i=\bm{\alpha}_i=(\bm{\alpha}_i^1,\ldots,\bm{\alpha}_i^{r})$.  When $\bm{x}_i\in\mathcal{S}_j$, we have 
\begin{equation}
\bm{\alpha}_i^j={\bm{U}^{j}}^\top\bm{x}_i={\bm{U}^{j}}^\top\bm{U}^{j}\bm{v}_i=\bm{v}_i.
\end{equation}
It follows from the assumption that
\begin{equation}
\max_p{\alpha_{pi}^j}>\mu.
\end{equation}
Let $\bm{b}_1=[\bm{b}_1^1;\ldots;\bm{b}_1^k]=-\mu \bm{1}$. Then $\bm{h}^j_i=\textup{ReLU}(\bm{\alpha}_i^j+\bm{b}_1^j)$ has at least one positive element.
On the other hand, since
\begin{equation}
\bm{\alpha}_i^l={\bm{U}^{l}}^\top\bm{x}_i={\bm{U}^{l}}^\top\bm{U}^{j}\bm{v}_i\quad  l\neq j,
\end{equation}
using the assumption of $\mu$, we have
\begin{equation}
\vert\alpha_{pi}^l\vert=  \vert{\bm{U}_{:p}^{l}}^\top\bm{U}^{j}\bm{v}_i\vert\leq\Vert{\bm{U}_{:p}^{l}}^\top\bm{U}^{j}\Vert\Vert\bm{v}_i\Vert\leq \mu,
\end{equation}
where we have used the fact $\Vert\bm{v}_i\Vert=1$ because $\Vert\bm{x}_i\Vert=1$.
It follows that $$\bm{h}^l_i=\textup{ReLU}(\bm{\alpha}_i^l+\bm{b}_1^l)=\bm{0}, \quad l\neq j.$$ 
Now we formulate $\bm{W}_2$ as 
\begin{equation}
\bm{W}_2=\left[
\begin{matrix}
\bm{q}_{11} &\bm{q}_{12}&\ldots &\bm{q}_{1k}\\
\bm{q}_{21} &\bm{q}_{22}&\ldots &\bm{q}_{2k}\\
\vdots &\vdots &\ddots&\vdots\\
\bm{q}_{k1} &\bm{q}_{k2}&\ldots &\bm{q}_{kk}\\
\end{matrix}
\right],
\end{equation}
where $\bm{q}_{lj}\in\mathbb{R}^{1\times r}$, $l,j=1,\ldots,k$. We have
$${z}_{ji}=\bm{q}_{j1}\bm{h}_i^1+\bm{q}_{j2}\bm{h}_i^2\cdots+\bm{q}_{jk}\bm{h}_i^k=\bm{q}_{jj}\bm{h}_i^j.$$
and 
$${z}_{li}=\bm{q}_{l1}\bm{h}_i^1+\bm{q}_{l2}\bm{h}_i^2\cdots+\bm{q}_{lk}\bm{h}_i^k=\bm{q}_{lj}\bm{h}_i^j.$$
Here we have let $\bm{b}_2=\bm{0}$.
Let $\bm{q}_{jj}\geq\bm{0}$ and $\bm{q}_{lj}=\bm{0}$, we have
$${z}_{ji}>{z}_{li}=0.$$
Therefore, if $\bm{x}_i\in\mathcal{S}_j$, we have ${z}_{ji}>0$ and ${z}_{li}=0$ $\forall 1\leq j\neq l\leq k$. Now performing k-means on $\bm{Z}=[\bm{z}_1,\ldots,\bm{z}_n]$ can identify the clusters trivially.

\end{proof}

\end{document}